\documentclass[acmsmall]{acmart}

\usepackage[T1]{fontenc}

\usepackage{mathtools}

\usepackage[nameinlink,capitalise]{cleveref}
\usepackage{multirow}
\usepackage{tabularx}
\DeclareMathOperator*{\argmax}{argmax}
\usepackage{xcolor}
\usepackage{graphicx}
\usepackage{tabularx}
\usepackage[noend]{algorithm2e}
\usepackage{setspace}
\usepackage[inline]{enumitem}

\crefname{line}{line}{lines}
\crefname{figure}{Fig.}{Figs.}
\Crefname{figure}{Fig.}{Figs.}
\crefname{equation}{Eq.}{Eqs.}
\Crefname{equation}{Eq.}{Eqs.}
\crefname{section}{Sec.}{Secs.}
\Crefname{section}{Sec.}{Secs.}
\crefname{definition}{Def.}{Defs.}
\Crefname{definition}{Def.}{Defs.}
\crefname{algorithm}{Alg.}{Algs.}
\Crefname{algorithm}{Alg.}{Algs.}
\Crefname{algocf}{Alg.}{Algs.}
\Crefname{appendix}{Appendix}{Appendices}

\usepackage{tikz}
\usetikzlibrary{arrows,matrix}

\newcommand{\omp}{ 
  \mathbin{
    \mathchoice
      {\buildcirclepm{\displaystyle     }{0.14ex}{0.95}{0.05ex}{.7}}
      {\buildcirclepm{\textstyle        }{0.14ex}{0.95}{0.05ex}{.7}}
      {\buildcirclepm{\scriptstyle      }{0.13ex}{0.955}{0.04ex}{.55}}
      {\buildcirclepm{\scriptscriptstyle}{0.08ex}{0.95}{0.03ex}{.45}}
  } 
}

\renewcommand{\omp}{ 
\tikz[baseline=(char.base)]{
\raisebox{0.6pt}{            \node[shape=circle,draw,inner sep=-0.3pt] (char) {{\footnotesize $\mp$}};}}
}

\usepackage{pbox}
\usepackage{array}

\newcommand{\PreserveBackslash}[1]{\let\temp=\\#1\let\\=\temp}
\newcolumntype{C}[1]{>{\PreserveBackslash\centering}p{#1}}
\newcolumntype{R}[1]{>{\PreserveBackslash\raggedleft}p{#1}}
\newcolumntype{L}[1]{>{\PreserveBackslash\raggedright}p{#1}}

\newcommand{\draft}[1]{{\color{gray}#1}}
\renewcommand{\draft}[1]{}

\graphicspath{{./images/}}
\begin{document}
%
%
%

\title[Falsification of Autonomous Systems in Rich Environments]{Falsification of Autonomous Systems in Rich Environments}

\author{Khen Elimelech}
\email{elimelech@rice.edu}
\affiliation{%
  \institution{Rice University}
  \city{Houston}
  \state{TX}
  \country{USA}
}
\author{Morteza Lahijanian}
\email{morteza.lahijanian@colorado.edu}
\affiliation{%
  \institution{University of Colorado Boulder}
  \city{Boulder}
  \state{CO}
  \country{USA}
}
\author{Lydia E. Kavraki}
\email{kavraki@rice.edu}
\affiliation{%
  \institution{Rice University}
  \city{Houston}
  \state{TX}
  \country{USA}
}
\author{Moshe Y. Vardi}
\email{vardi@rice.edu}
\affiliation{%
  \institution{Rice University}
  \city{Houston}
  \state{TX}
  \country{USA}
}

{\let\thefootnote\relax\footnote{{Work on this paper was supported by Office of Naval Research (ONR)
MURI grant no. N00014-20-1-2787.}}}

\begin{abstract}
Validating the behavior of autonomous Cyber-Physical Systems (CPS) and Artificial Intelligence (AI) agents, which rely on automated controllers, is an objective of great importance. In recent years, Neural-Network (NN) controllers have been demonstrating great promise, and experiencing tremendous popularity. Unfortunately, such learned controllers are often not certified and can cause the system to suffer from unpredictable or unsafe behavior. To mitigate this issue, a great effort has been dedicated to automated verification of systems. Specifically, works in the category of ``black-box testing'' rely on repeated system simulations to find a falsifying counterexample of a system run that violates a specification. As running high-fidelity simulations is computationally demanding, the goal of falsification approaches is to minimize the simulation effort (NN inference queries) needed to return a falsifying example. This often proves to be a great challenge, especially when the tested controller is well-trained. This work contributes a novel falsification approach for autonomous systems under formal specification operating in uncertain environments. We are especially interested in CPS operating in rich, semantically-defined, open environments, which yield high-dimensional, simulation-dependent sensor observations as inputs to the controller. Our approach introduces a novel reformulation of the falsification problem as the problem of planning a trajectory for a ``meta-system,'' which wraps and encapsulates the examined system; we call this approach: meta-planning. The search technique makes minimal assumptions on the system, and poses no limitation on the specification, environment parameters, or controller, which is treated as a black-box. The approach results in testing less inputs until finding a falsifying example, compared to serial input sampling. It also avoids redundant calculations and requires less effort for each test case, by invoking only incremental updates to the autonomous system's trajectory at each iteration, using partial simulations. This formulation can be solved with standard sampling-based motion-planning techniques (like RRT) and can gradually integrate domain knowledge to improve the search, based on its availability, and can even work with no domain knowledge at all. We support the suggested approach with an experimental study on falsification of an obstacle-avoiding autonomous car with a NN controller, where meta-planning demonstrates superior performance over alternative approaches.
\end{abstract}

\maketitle

\newcommand{\prop}{\textit{prop}}
\newcommand{\comp}{\textit{components}}

\newcommand{\status}{\mathtt{status}}

\renewcommand{\state}{\xi}
\newcommand{\statetraj}{\overline{\state}}
\newcommand{\statespace}{\Xi}
\newcommand{\scene}{s}
\newcommand{\scenespace}{S}
\newcommand{\scenetraj}{\overline{\scene}}
\newcommand{\scenetrajspacece}{\overline{\scenespace}}
\newcommand{\env}{\textit{env}}
\newcommand{\envspace}{\textit{Env}}

\newcommand{\obs}{z}
\newcommand{\obsspace}{Z}
\newcommand{\obstraj}{\overline{\obs}}

    \section{Introduction}

\subsection{Background and motivation}
As autonomous Cyber-Physical Systems (CPS) and Artificial Intelligence (AI) agents become embedded in various aspects of modern life, the goal of certifying the behavior and ensuring trustworthiness of such systems is becoming one of great importance \cite{Seshia2022VerifiedArtificial}.
To operate autonomously, such systems and agents often rely on automated controllers, which are designed to translate a stream of sensor observations or system states into a stream of commands (controls) to execute, in order to maintain a safe behavior, or robustly perform a specified task.

Traditionally, controllers had to be expertly designed, e.g., by meticulously considering physical and mechanical aspects of the system. In recent years, however, computational Neural-Network (NN) controllers have been experiencing tremendous popularity. These can handle complex, high-dimensional sensor observations, such as images, and enable effective control of highly-complex dynamical systems, such as racing cars, snake robots, high Degree-of-Freedom (DoF) manipulators, and dexterous robot hands, which have been a great challenge in the controls and robotics communities.
Such controllers are typically built (``trained'') by compressing numerous examples (``training data'') using statistical machine learning techniques, in an attempt to yield a certain behavior. Common techniques include Reinforcement Learning (RL) \cite{Singh2022ReinforcementLearning}, from repeated trial-and-error control attempts, until apparent convergence to a desired behavior, and Imitation Learning \cite{Zare2024SurveyImitation}, from demonstrations of either a human operator or a traditional controller. 
Unfortunately, such learning methods generally do not provide a guarantee that the resulting controller will robustly exhibit the desired behavior; hence, relying on these controllers can cause the system to suffer from unpredictable or unsafe behavior on edge cases.
While there has been a recent efforts to advance controller synthesis \cite{Kress-Gazit2018SynthesisRobots,Raman2015ReactiveSynthesis,Wells2021FiniteHorizonSynthesis}---that is, the automated creation of controllers that are guaranteed to comply to given specification by design---these usually fail to scale beyond simple scenarios; and, more importantly, are only certified in relation to the assumed (and often simplified) system models.

\subsection{Problem positioning and definition}
To mitigate the aforementioned issues, a great effort has been dedicated to automated verification \cite{Deshmukh2019FormalTechniques} of CPSs, which, as we will discuss, has proven to be a formidable challenge; this work comes to contribute to this effort.
A system \cite{Lee2011StructureInterpretation}, in technical view, is a machine that maps an input signal to an output signal. For verification, we wish to examine how the variability of the possible inputs may affect the system output, which should be tested against a formal specification, indicating a desired behavior or safety requirements. Specifications may be given in various formats, e.g., using automatons, state predicates, or temporal logic \cite{Pnueli1977TemporalLogic}.
We are especially interested in \emph{autonomous} systems, in which we care to test the automated system controller, for which the inputs are the stream of sensor observations and/or the initial conditions. This comes in contrast to controllable systems, where we want to test the system behavior for different streams of control commands as inputs.
We should also mention that while there has been a considerable effort to develop techniques for robustness verification for NN controllers or AI-based system components in isolation \cite{Katz2017ReluplexEfficient,Ehlers2017FormalVerification,Wang2018EfficientFormal,Weng2018FastComputation}; yet, these approaches are inadequate in cases like ours, when the controller inputs and outputs are a part of the closed-loop autonomous system, and the specification is defined in the system-level.

Works on automated verification of systems \cite{Deshmukh2019FormalTechniques} can broadly be divided into two categories.
Works in the first category rely on reachability analysis, in order to answer the question ``can the system end up in an unsafe state?'' Since such analysis requires knowledge on the system dynamics, it is often referred to as ``white-box verification.''
Prominently, many such techniques for verification of traditional continuous controllers rely on Hamilton-Jacobi reachability analysis \cite{Bansal2017HamiltonJacobiReachability,Mitchell2005TimedependentHamiltonJacobi}. While it is more challenging for NN controllers, due the deep non-linearity and numerous parameters, some recent advances showed that such analysis can be performed for NN controllers with known, exploitable architecture and/or low-dimensional inputs \cite{Akintunde2018ReachabilityAnalysis,Ivanov2019VerisigVerifying,Ivanov2020VerifyingSafety,Xiang2018ReachabilityAnalysis}.
Nevertheless, as the reachability problem is generally undecidable \cite{Henzinger1998WhatDecidable}, this type of verification is limited to relatively simple systems or systems under specific constraints. Such techniques are also not natively suitable for temporal specifications, which are defined over state trajectories. 
Works in the second category rely on performing repeated simulations of a system run, in an attempt to find a counterexample, which would \emph{falsify} the controller, i.e., an example of a system run, under some conditions, that violates the specification. 
Since this type of analysis typically does not require knowledge on the system dynamics, but only examination of input-output pairs, it is often referred to as ``black-box verification'' or ``black-box testing.''
Various falsification approaches suggest strategies for efficiently choosing promising test cases.
As running high-fidelity simulations is often computationally demanding, the goal (and metric of success) of such approaches is to minimize the total simulation effort needed to return a falsifying example; when considering a NN-controller, this goal may correspond to minimizing the number of control loops (NN inference passes) needed to find such an example. This often proves to be a great challenge, especially when the controller is well-trained.

This paper suggests a novel falsification approach for autonomous systems under formal specification operating in \emph{uncertain environments}. We are especially interested in CPS operating in \emph{rich, semantically-defined, open environments}, which may yield \emph{high-dimensional sensor observations} as inputs to the controller---as is the case in sensor-based autonomous driving. Our approach aims at (i)~minimizing the number of simulation effort needed to find a falsifying example, (ii)~being applicable to the most general scenarios, and (iii)~making minimal assumptions on the availability of domain knowledge.
While we are especially interested in systems with NN controllers, this work is, in fact, relevant to verification of general controllers, which are treated as black-boxes.

\subsection{Falsification: a review}
\label{sec:related-work}
To properly explain the novelty of our approach in relation to existing solutions, before introducing it, we first provide a summary of the relevant literature. A knowledgeable reader may choose to skip this review directly to \cref{sec:intro-approach}.
Also, while there is some distinction between testing of autonomous and controlled systems, the shared assumption that the input space can be parameterized using a discrete, finite, and samplable set of points makes most falsification techniques applicable to either problem. We will therefore ignore the distinction between the two cases for the rest of this review.

\subsubsection{Robustness optimization}
These days, the most prominent approach to falsification \cite{Corso2021SurveyAlgorithms}, when the specification is provided as a temporal logic formula over the system trajectory, is reformulating it into an optimization problem of the \emph{signal robustness} \cite{Fainekos2009RobustnessTemporal,Akazaki2015TimeRobustness}---a scalar measure of how ``close'' the output signal (system trajectory) is to failing the specification.
The notion of robustness, often referred to as \emph{quantitative semantics}, is generally limited to continuous signals and temporal logics based on continuous parameters, most prominently Metric Temporal Logic (MTL) \cite{Koymans1990SpecifyingRealtime} and Signal Temporal Logic (STL) \cite{Maler2004MonitoringTemporal}. Intuitively, for a specification-satisfying signal, the robustness function returns the radius of the largest cylinder around this signal, in which all contained signals also satisfy the specification; or, in other words, it measures how much we can perturb this signal before the specification are compromised.
With that, the falsification problem can be posed as an optimization problem, in which we search for the trajectory of minimal robustness. We note that the robustness function requires domain knowledge and may be non-trivial to calculate in practice.

Since evaluation of the system with a given input requires performing a simulation run, the optimization process is most often based on discrete, serial sampling. The user starts by sampling a random input, to be fed into a simulator, which then generates a system trajectory to be tested (imposing a time limit, if needed). The process then repeats, until a falsifying example is found.

\subsubsection{Optimization techniques}
Various techniques for robustness optimization differ in the way they guide the input sampling (search) process and can largely be divided into ``active'' techniques and ``passive'' techniques \cite{Ramezani2022OptimizationbasedFalsification}.

In ``active'' optimization techniques, dominated by Bayesian optimization \cite{Deshmukh2017TestingCyberPhysical}, one tries to learn a (probabilistic) model of the input-to-robustness function, to select the next-best signal to test. These techniques often rely on a Gaussian Process (GP) in order to learn this model, and require making numerous assumptions on the underlying process distributions. While some common assumptions seem to work well in many cases, these are often not grounded in actual knowledge, but are chosen to ease of the computation. Bad choices might hinder convergence---something that is regardless only guaranteed for mappings. 
As shown, e.g. in \cite{Ghosh2018VerifyingControllers}, the optimization performance can drop significantly when dealing with complex or composite specification, which might compromise the smoothness assumptions. These cases thus require usage of dedicatedly-crafted optimizers that use expert knowledge to effectively decompose the specification (e.g., \cite{Mathesen2021EfficientOptimizationBased,Zhang2019MultiarmedBandits}).
While it is potentially possible to bias active techniques with domain knowledge, this knowledge is typically expected as prior sampling distributions (as in \cite{Ramezani2023FalsificationCyberPhysical}), which is often not be trivial to achieve and unavailable.

``Passive'' optimization techniques, often referred to as ``direct search'' or ``random search'' methods, rely on various non-learning techniques to ensure the sampling procedure of the inputs well-covers the space.
These include, for example, straightforward sampling approaches like Line-Search \cite{Ramezani2022TestingCyber} and Simulated Annealing \cite{Abbas2013ProbabilisticTemporal,Aerts2018TemporalLogic}, or more advanced ones, like the Cross Entropy Method \cite{Sankaranarayanan2012FalsificationTemporal,Kim2016ImprovingAircraft}---a type of importance sampling in which one iteratively estimates a sampling distribution based on the output values (but does not estimate a model of the system); this approach, like Bayesian optimization, also require setting assumptions on the underlying distributions.
Notably, all of the aforementioned optimization techniques assume the input space is defined in a continuous box, and do not natively support discrete input variables.
Although seemingly less popular today, evolutionary (genetic) algorithms have also been examined as a direct search technique \cite{Zhao2003GeneratingTest,Hansen1996AdaptingArbitrary}. In such a serial approach, one starts from a ``population'' of $n\geq1$ samples (rather than a single one, in the previous approaches); then, in each step, a new generation of $n$ samples is created, through crossovers (mergers) and mutation among the previous population.
These evolutionary methods precede the notion of robustness, and can be used to optimize a general objective function (provided by the user as domain knowledge); they may also generally support discrete variables.

It is worth mentioning that several recent works \cite{Zhang2023FalsifAIFalsification,Dreossi2019CompositionalFalsification}, focus specifically on robustness optimization for systems with NN components. These works suggest various ways to use feedback from monitoring of the NN inference process/outputs, in order to guide future sampling and the overall optimization process. We do not assume access to such expert knowledge.

\subsubsection{Planning-based falsification}
\label{sec:intro-planning}
Beyond the optimization approach covered, another prominent approach for solving the falsification problem is by formulating it as a path/motion planning, and employing sampling-based motion planning algorithms \cite{Orthey2024SamplingBasedMotion,Elbanhawi2014SamplingBasedRobot} to solve it. Such algorithms are celebrated in robotics research for their ability to effectively explore high-dimensional state space (e.g., configuration spaces of high-DoF manipulators).
In contrast to the aforementioned optimization-based techniques, in which one samples complete inputs for the system and use them to generate a full system trajectory to test, these approaches (e.g.,  \cite{Cheng2008SamplingbasedFalsification,Aineto2023FalsificationCyberPhysical,Plaku2009HybridSystems,Plaku2013FalsificationLTL,Esposito2005AdaptiveRRTs,Tuncali2019RapidlyexploringRandom,Ernst2021FalsificationHybrid})  build output system trajectories incrementally, by interleaving sampling of local disturbances as partial system inputs (e.g., discrete control inputs, system noise, or actions of other agents) and performing of local, time-bounded system simulation. 
They treat the injection of disturbances as a mean to control the system trajectory, and use planning techniques to effectively grow a tree of diverging trajectories in hope to find a falsifying one. The specification validity is checked on each trajectory prefix, after each extension, until a trajectory with violation is detected.
These approaches are thus more relevant to controllable hybrid-systems or reactive systems with dynamic disturbances, and not, e.g., to verify robustness to varying environments/initial conditions. They are also, by their nature of operation, limited in the type of specification they can falsify: they are designed to handle ``safety specifications,'' in which the system is safe by default until at some point (after incurring a certain disturbance) it becomes unsafe; they are not designed to handle ``liveness specifications''  \cite{Lamport1977ProvingCorrectness},  in which a partial system trajectory does not satisfy the task, until at some point it does. 
The latter case is especially relevant for controllers of robotic systems performing abstract tasks, defined, e.g., as symbolic reach-avoid constraints, PDDL goals \cite{Fox2003PDDL2Extension}, or finite-LTL formulae~\cite{DeGiacomo2013LinearTemporal}.

Other approaches that in a similar fashion incrementally build a diverging system trajectory tree were also examined, including Monte Carlo Tree Search (MCTS) \cite{Zhang2018TwoLayeredFalsification}, a reward-based tree growth technique, and (deep) RL \cite{Julian2020ValidationImagebased,Yamagata2021FalsificationCyberPhysical}, which uses learning to generate a policy to guide the tree growth.

\subsubsection{Parameterization of the input space}
\label{sec:review-parameter}
As mentioned in the beginning of this review, to automatically generate test inputs for the system, falsification techniques must consider some sort of finite input parametarization. It is ubiquitous to assume the values of these parameters are limited to continuous ranges (``boxes'') \cite{Zhao2003GeneratingTest,Sankaranarayanan2012FalsificationTemporal,Ramezani2022TestingCyber,Deshmukh2017TestingCyberPhysical}.
In optimization-based approaches, the number of parameters must also be predefined and set.
Further, most existing techniques and tools assume the test input to be a signal that can be presented as a parameterized curve \cite{Ernst2021ARCHCOMP2021,Zhao2003GeneratingTest,Sankaranarayanan2012FalsificationTemporal,Deshmukh2017TestingCyberPhysical} or a discrete sequence of disturbances \cite{Corso2021SurveyAlgorithms}, and that these parameters are externally controlled, and thus can be sampled arbitrarily \cite{Annpureddy2011STaLiRoTool,Donze2010BreachToolbox}. 
While these assumptions might be reasonable for testing controllable systems, where the control-signal is often of a numeric, low-dimensional, free-to-choose vector, they are less so for autonomous systems, where a realistic sensor-signal is often high-dimensional and cannot be sampled validly without simulating the system (due to its continuous dependence on the state).
As a result, making such assumptions imposes severe limitation on the systems we can analyze and techniques we can use.
Indeed, existing works on falsification of autonomous systems typically consider very simple scenarios, such as an adaptive cruise control systems \cite{Zhang2023FalsifAIFalsification}, a multi-aircraft conflict-resolution system \cite{Plaku2009HybridSystems}, an Abstract Fuel Control (AFC) system \cite{Jin2014PowertrainControl}, an insulin infusion system \cite{Sankaranarayanan2012FalsificationTemporal}, or a goal-reaching navigation system \cite{Ghosh2018VerifyingControllers}. In these systems the inputs are simple: initial conditions represent the initial system state, and the input signal describe \emph{observations of an external quantity}, given as a continuous, low-dimensional signals, which can fluctuate freely, e.g., in the mentioned examples, the position of the leading car, the direction of wind, the engine speed, the blood glucose level, or added motion noise.

A few recent examples \cite{Julian2020ValidationImagebased,Sankaranarayanan2017ModelbasedFalsification,Dreossi2019VerifAIToolkit,Seshia2022VerifiedArtificial} mitigated some of these issues by modeling the test input as a set of rich initial conditions, or a description of ``an environment,'' used to initiate a simulation, which could then generate the sensor signal (instead of assuming it can be sampled directly). We support this concept and find it to be more appropriate for general autonomous systems.

\subsection{Our approach: falsification as meta-planning}
\label{sec:intro-approach}

\subsubsection{Standardized environment-based testing model}
As mentioned, instead of trying to reason over sensor-observation signals as the test inputs for our autonomous system, we believe a more general and more appropriate model should consider \emph{environments} as inputs.
Thus, as the first part of our contribution, we formulate and standardize a testing model that considers an environment as input (as illustrated in \cref{fig:systems}). An environment is a description of all parameters \emph{external} to the system that can affect the system throughout its run or the specification. 
The formalism we suggest extends the concept of rich initial conditions and is specifically intended to support systems operating in and observing rich, modular, open-world environments. The environment, together with the state of the system in it, is what would determine the observation stream, which would be generated incrementally through a simulation.
The environment-based model may be viewed as an alternative input paramaterization to the signal-based standard. Accordingly, it may be utilized with the previously-covered falsification techniques, which would enable their usage for autonomous systems---something that is not natively supported when considering the the standard model; we demonstrate this aspect in our experimental results.

\subsubsection{Falsification as meta-planning} 
As the main contribution of this paper, and utilizing the suggested environment-based input parameterization, we propose a new direct-search falsification approach for autonomous systems under general formal specification.
In our approach, we reformulate the falsification problem, the search for inputs that would fail the system, as the problem of planning a trajectory for an (under-actuated) meta-system, which wraps and encapsulates the examined system; we call this approach: meta-planning. \emph{Each meta-state encapsulates both a system input (environment) and its corresponding output (a simulated system trajectory in the environment)}.
A meta-control, which can be applied on meta-states, invokes a local change to the input and a corresponding update to the system trajectory. The goal meta-region contains all the meta-states in which the system trajectory indicates failure of the specification. Note that this goal definition abstracts away the actual specification and is a ``classical'' goal predicate, even if the specifications are temporal. This allows us to apply standard and efficient planning techniques to search the meta-state space and discover a meta-trajectory that, starting from some initial guess, reaches the a goal meta-state and solves the problem. 
Note that the objective of such planning is simply to find a goal meta-state, corresponding to a falsifying example; the meta-trajectory, which represents the progression of the falsification process from an initial example into a falsifying one, through a sequence of environment mutations, is of lesser importance.

We should clarify that this contribution is not a new search algorithm, but a \emph{reformulation of the falsification problem,} which enables its efficient solution using existing tools. This idea is equivalent to the idea of formulating the problem as robustness optimization, which could then be solved with a variety of optimization techniques, as we covered.
On that regard, it is important to note that optimization-based formulation may not always be possible, if there is not a well-defined value function, e.g., when considering non-quantitative specifications.
Our less-restrictive formulation does not force the falsification problem into an optimization format (and specifically not robustness optimization), but maintains its inherent definition as a search problem, whose solution can, as we shall explain, be guided by a heuristic, if one is available. In general, this formulation does not require any domain knowledge to be used; with that said, it can incorporate and benefit from a variety of domain knowledge sources, if they are available (as we show in \cref{sec:solving}).

This general falsification technique is, in fact, applicable to falsification of any black-box system, beyond merely autonomous systems, as it only requires input-output knowledge. Yet, as an important feature, when considering our particular case of an autonomous system with an automated controller, we show that this technique can inherently benefit from efficient, incremental simulation, to minimize redundant calculations for each test case. 

\subsubsection{Summary of contributions}
To summarize, this paper contributes
\begin{enumerate*}[label=(\roman*)]
\item a standardized testing model for autonomous systems operating in a rich, open environment (\cref{sec:model}), in support of \linebreak
\item a novel formulation of the falsification problem as a meta-planning problem, (\cref{sec:approach}). We also provide 
\item a practical explanation and guidelines on how to solve the meta-planning problem using sampling-based motion planning algorithms (namely, RRT), and how to potentially incorporate domain knowledge to accelerate the search  (\cref{sec:solving}). Finally, we present
\item an experimental study on using meta-planning for falsification of an obstacle-avoiding autonomous car with a NN~controller, including a comparison against alternative falsification approaches, in which meta-planning shows superior performance (\cref{sec:evaluation}). 
\end{enumerate*}
Pseudocode summary of our algorithms is provided in Appendix~\ref{appen:algorithms}, and extended theoretical discussion and comparison is provided in Appendix~\ref{appen:theory}.
    
    \section{Standardizing the testing model: autonomous systems in uncertain environments}
\label{sec:model}
We care to examine an simulatable autonomous dynamical system, referred to as the ``system," acting in an \emph{environment}, which encodes the external factors affecting the system's operation.

\subsection{Environment formalism}
\label{sec:envs} 
An \emph{environment-type} defines the set of properties or variables one should specify in order to describe the surroundings in which the system can operate. An object of the environment-type, which determines a specific valuation to these variables, is referred to as an \emph{environment-state}. An~environment-state can be observed by the system, and potentially changed by it.

Variables can be classified as either environment \emph{parameters} or \emph{elements}---a separation we will later exploit for computational gains.
Simply put, parameters describe (i)~essential information, which (ii)~affects the system's observation and/or dynamics globally, regardless of the system state; elements describe (i)~optional, additive features used to enrich the environment, which (ii)~may be observed locally, from only a subset of system states (making the environment partially-observable by the system).
The environment-type defines for both parameters are elements their type, e.g., a scalar, a vector, a function, or even a time-dependent model. 
By convention, elements should be organized into \emph{collections}, grouped together by the features they describe; such collections may be ordered (i.e., a vector of elements) or unordered (i.e., a set of elements), and may also indicate the minimal/maximal number of elements that may be provided in each environment-state, or other restrictions, e.g., on the mutual-exclusiveness of the elements.

This formalism we suggest is not limited to continuous variables in box-ranges, but can include discrete and symbolic variables, and even be used to encode time-varying perturbations. As we use collections to specify the environment elements, the number of variables in each environment-state is not predetermined---allowing us to model systems operating in rich, open environments. 
We provide a concrete example at the end of this section.

\subsection{System model}
While we support both continuous and discrete systems, for generality, we assume the system operates under continuous dynamics, modeled as:
\begin{equation}
    \begin{split}
        \dot{\state} &= f(\state,u,\env), \\
        u &= g(\obs), \\
        \obs &= h(\state,\env),
    \end{split}
\end{equation}
where $\state \in \statespace$ is the system state, $\env\in \envspace$ is the environment-state of environment-type $\envspace$, $u \in U$ is a control action, $f: \statespace \times U \times \envspace \to \statespace$ is the vector field, $g: \obsspace \to U$ is a ``black-box'' NN-controller, and $h: \statespace \times \envspace \to \obsspace$ is the system's observation (sensor) model. 
Note that while the environment-state can encode time-variant parameters, its definition is assumed to be static throughout the system operation.
We use $\state_t$ to mark the system state at time $t\in[0,T]$, and $\statetraj_{0:T}$ to mark the concatenation of states from time~$0$ to time~$T$, a time-parameterized continuous trajectory. If the length of the trajectory is not of interest, we may drop the subscript notation.

\begin{figure}[ht]
\centering

{%
\setlength{\fboxsep}{0pt}%
\setlength{\fboxrule}{2pt}%
\fbox{\includegraphics[width=0.7\textwidth]{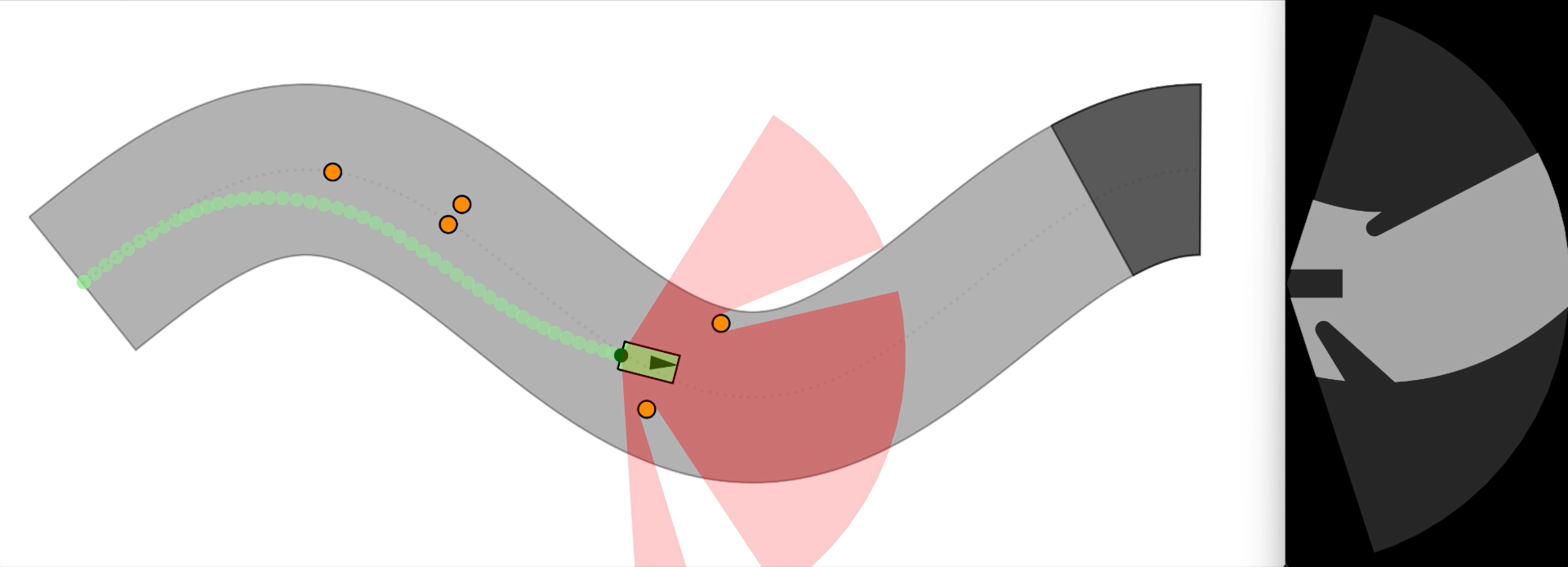}}
}%

\caption{Our running example: an autonomous car acting in an ``obstructed track'' environment, whose \emph{parameters} define the track shape, while its \emph{elements} are the obstacles, which are organized in the ``obstacle collection.'' The car trajectory is in green. At each state, the car observes the track using a lidar sensor; the observation image of the area highlighted in red is shown on the right. The car NN-controller is trained to take in the stream observation images and steer the car to the end of the track while avoiding collision.}
\label{fig:car}
\end{figure}

Note that, \textbf{for conciseness of discussion, we will consider the initial system state $\state_0$ to be given and set}, and the system to contain no actuation nor sensing noise, making the environment-state the only controllable simulation input, and the environment-state variability the only source of uncertainty. These assumptions are not essential to our approach. Further, we will assume the control input is updated at some constant frequency and that the control execution is free of noise; though, again, this is not a requirement.
Also, note that from now on, wherever the context is clear, we will simply use ``environment" to refer to an environment-state.

\subsection{Task specification}
We assume the NN-controller was trained, e.g., using RL, to guide the system towards completing a \emph{task}, while being robust to environment perturbation; we make no assumptions on the NN or the way it was trained.
Unlike standard approaches, the task may be formally specified as a constraint over the system trajectory \emph{and the environment}, e.g., defined using a temporal logic formula. \linebreak We do not limit the type of logic, and support both safety-type and liveness-type specifications.
To evaluate the system's success in the task, we assume availability of a ``status'' predicate:
\begin{equation}
    \status(\statetraj,\env) \in \{0,1\},
\end{equation}
which encapsulates evaluation of the task constraint, and abstracts away the specification's specifics. Evaluation of this predicate can be done with an appropriate model checker.
For a trajectory $\statetraj$, the outcome of a system run in an environment $\env$ over a period of time, $\status$ returns $1$, if the trajectory satisfies the task specification, or $0$, if it does not.

\subsection{Controller falsification: problem definition}
As means to verify the behavior of the system and the controller's robustness, our objective is to try and to falsify the specification. Meaning, we wish to find and return a witness of failure, i.e., an example of an environment $\env$ and a system trajectory in it, for which the task specification is violated, if such a witness exists.
While we assume we can easily observe whether the system trajectory conveys a success or failure of the task, if the controller is well-trained, it might not not be trivial to find an scenario in which the system fails.
Thus, we specifically care to falsify the controller using minimal simulation effort, which we measure using the number of (inference) queries to the NN controller. Further, we would like to rely on a system-agnostic falsification technique, which minimizes the reliance of expert knowledge.

\subsection{A running example: an autonomous car}
\label{sec:running-example}
While the ideas presented in the following sections are relevant to a general system, to ground the discussion, we will consider a running example of ``an autonomous car on an obstructed track,'' as depicted in \cref{fig:car}.

For a environment of type ``obstructed track,'' the variables are: track curve, track range, track width, and a collection of obstacles. The first three components are environment parameters of types $f\colon\, \mathbb{R}\rightarrow\mathbb{R}$, $[x_{\text{start}},x_{\text{end}}]\in\mathbb{R}^2$, and a non-negative scalar $\in\mathbb{R}^+$, respectively; the fourth component is a set of ``circular obstacle" elements, each of which of type $[x,y,r]\in\mathbb{R}^2\times\mathbb{R}^+$. Each environment defines several regions in the $xy$ plane (over which the track is laid), including the track end zone, the obstructed area (the area covered by the obstacles), and the track shoulders (the area beyond the edge of the track).

The state $\state\doteq[x,y,\phi,\alpha,v]^T\in\statespace$ defines the car's origin position and heading (from which we can derive the location of the car's overall shape), its steering angle, and speed. The observation model $h$ returns a lidar scan of the track from the car's pose (an image). The controller $g$ sets the acceleration and steering velocity, given the stream of lidar scans (and the current steering angle), and $f$ complies to a bicycle model.
The controller was trained to steer the car to the end of the track while avoiding collision, regardless of the track curve and obstacle placement. 
This task can be defined with the finite LTL (LTLf) \cite{DeGiacomo2013LinearTemporal} formula
\begin{equation}
\diamondsuit (\state.\textit{shape} \subseteq \env.\textit{end\_region}) \wedge \square \neg(\state.\textit{shape} \cap \env.\textit{obstructed\_area}) \wedge \square \neg(\state.\textit{shape} \cap \env.\textit{shoulders}).
\end{equation}

Note that since the car state (signal) is composite and is not defined in the same space in which the environment regions (which define that atomic propositions) are defined (the $xy$ plane), the notion of robustness \cite{Fainekos2009RobustnessTemporal} is not well-defined for this scenario.
In any case, we can define for this task a $\status$ predicate, which returns $1$ if the car's trajectory satisfies the formula, and $0$ otherwise.

Our goal in this scenario is to efficiently find an obstructed track environment in which running the autonomous car would result in collision.

\begin{figure}[b]
\centering
\scriptsize
\includegraphics[scale=0.3]{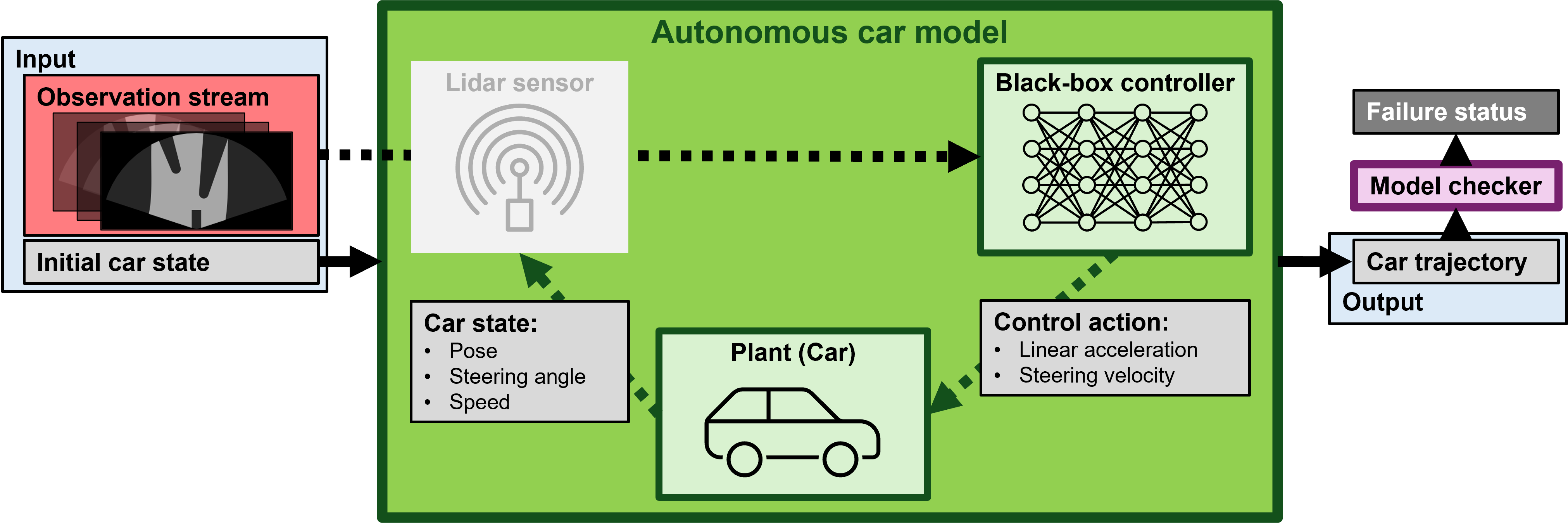}\hspace{15pt}$ $

(a) A standard-yet-inappropriate testing model for the autonomous system, attempting to model the sensor observation stream as input.
\vspace{\baselineskip}

\includegraphics[scale=0.3]{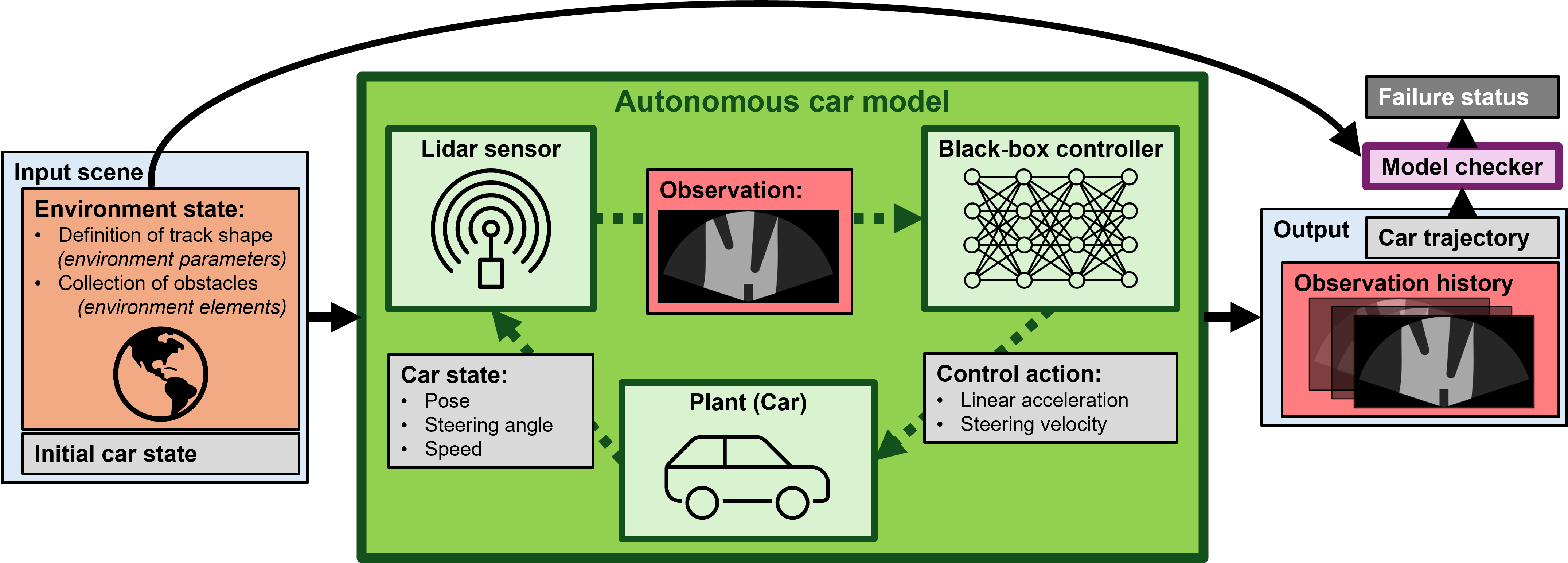}

(b) Our suggested amended testing model, considering a scene as input, and the sensor observation stream as the simulation output.

\caption{Testing models used for falsification of autonomous car system.}
\label{fig:systems}
\end{figure}

\subsection{The testing model}
As previously covered, in typically-examined systems, the observation trajectory is assumed to be externally-controllable.
Hence, the standard testing model used for falsification is based on paramaterization of the sensor signal as a test input.
Yet, here, we do not make this assumption, but consider the sensor observation to be a function of the environment-state and the system state.
This means that to generate a temporally-consistent sensor signal, it must be generated incrementally by running a simulation of the autonomous system, at the same time the system trajectory is generated. This property is amplified when considering high-fidelity observations, like a camera stream, which can only be generated by the simulator.
Accordingly, the standard testing model, in which the sensor signal is considered an input, is simply inappropriate for testing our scenario, as depicted in \cref{fig:systems}a.

To mitigate this gap, we suggest a slightly-but-crucially amended model, based on our environment formalism, as presented in \cref{fig:systems}b. 
In this model, the simulation and test input is defined by an environment-state $\env$ and an initial system state $\state_0$, which together comprise a \emph{scene}\footnote{This terminology is consistent with standard falsification tools like Scenic \cite{Fremont2023ScenicLanguage}.}. The simulation output is the system trajectory $\statetraj_{0:T}$, alongside the $\obstraj_{0:T}$ sensor observation history, which are then fed to a model checker, to generate the test result. As we shall see, including explicitly the observation history as a simulation output will later allow us to exploit incremental simulation when analyzing the system.
This model also allows us to support more general task specifications, which may also depend on the environment-state, on top of the system trajectory.

    \section{Main contribution: reformulating the falsification problem as meta-planning}
\label{sec:approach}
Our goal in falsification is to search the input space for one for which the corresponding output indicates specification failure.
In the case of an autonomous system, as we previously modeled, this means searching the scene space, i.e., all possible initial states $\times$ all possible environment-states of the relevant environment-type, for a scene in which the system would not satisfy its task. 
Instead of serially sampling inputs and performing independent tests, we suggest to start from an initial test, and search by performing sequences of gradual, local changes to it, until one of these sequences leads us to a test that satisfies the condition. As we introduce ahead, this approach to falsification can be formulated as planning a path to ``meta-system,'' which encapsulates the examined system.

\subsection{Meta-planning in the meta-state space}
We mentioned that to solve the falsification problem, we would like to search the scene (input) space.
Yet, we note that the goal of the search is defined over the system trajectory (output). By such, for each input that we reach in our search, we must also simulate the corresponding output, in order to validate its status or measure our search progress. This means, in fact, that when searching the input space, we implicitly search the composite space of of inputs $\times$ outputs.
We suggest here to to reason about and search this composite space \emph{explicitly}, as this could allow us to better guide our search---by analyzing the diversity of both the sampled environments (like optimization-based approaches do) and the system reactions (like planning-based approaches do).
In our case, this suggestion means explicitly reasoning about and search the composite space $\scenetrajspacece \subset$ environments $\times$ system-trajectories $\times$ observation-histories, containing all valid \emph{simulated scenes}, where a \emph{simulated scene} $\scenetraj \doteq (\env, \statetraj_{0:T}, \obstraj_{0:T})$ is said to be \emph{valid}, if the system trajectory $\statetraj_{0:T}$ and observation history $\obstraj_{0:T}$ are consistent with the system's observation, controller, and transition models, and environment~$\env$. In other words, the simulated scene is valid if running the system in this scene produces (or can produce, for non-deterministic scenarios) the given system trajectory and observation history. The space $\scenetrajspacece$ shall serve as the ``state space'' in our planning-based formulation. Crucially, since we do not plan a trajectory for the system, but plan in the space of system trajectories, we refer to this problem as \emph{meta-planning}, to simulated scenes as \emph{meta-states}, and to $\scenetrajspacece$ as the \emph{meta-state space}.\linebreak
The task status predicate can be restated accordingly as a predicate on meta-states:
\begin{equation}
    \status(\scenetraj) \in [0,1].
\end{equation}

\begin{figure}[b]
\centering
\includegraphics[width=\textwidth]{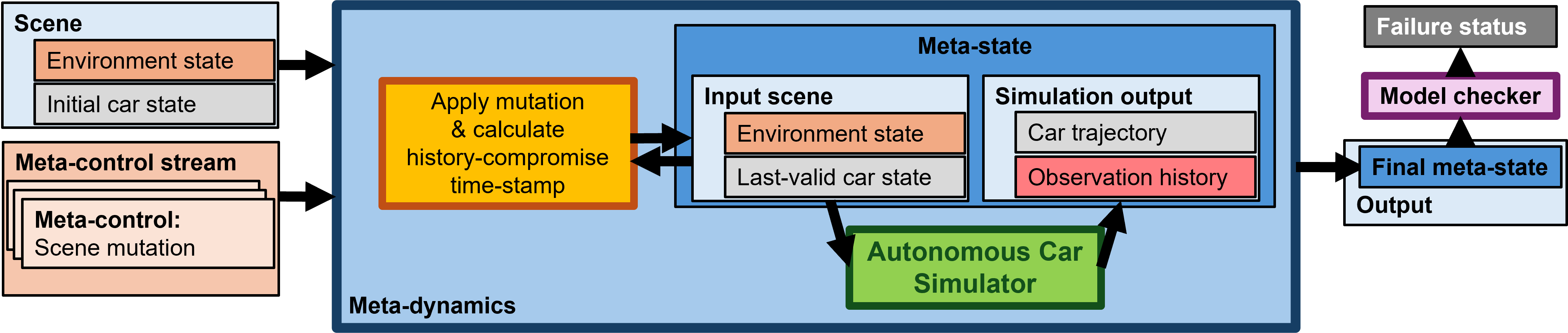}
\caption{The meta-system model.}
\label{fig:meta-system}
\end{figure}

\subsection{Meta-control}
\label{sec:meta-control}
To transition between meta-states, we need to define meta-control actions.
Each meta-control action corresponds to a scene mutation, followed by a simulation of a system run in the mutated scene, in order to update the system trajectory and determine the new meta-state.
Note that since we consider here the initial state to be set, we may more simply consider only mutating the environment, as we shall do from now on. In the general case, mutation may also change the initial state.

\subsubsection{Environment mutation}
An environment mutation is a transition from one environment-state to another, in the environment-space induced by an environment type (e.g., all obstructed tracks). \textbf{For the interests of this paper, we restrict the mutation only to changes to the environment-element collections,} excluding the environment parameters (and the initial state).
In our running example, a legal mutation would convey a change in the set of obstacles placed on the track, without changing the track itself.

To specify a mutation, we use a 3-tuple of the form $m\doteq(coll,op,elements)$ where ${coll}$ is the name of the element collection, $op$ is the operation to be performed on this collection, and $elements$ are inputs to this operation.
We consider two basic operations: $\oplus$ and $\ominus$, which indicate \emph{addition} and \emph{subtraction} of elements to/from a collection, respectively.
The result of applying the mutation $({coll},\oplus,elements)$ (respectively, $({coll},\ominus,elements)$) on an environment $\env$ is a new environment, marked $\env \oplus_{coll} elements$ ($\env \ominus_{coll} elements$), in which the value the collection ${coll}$ is $\env.{coll}\cup elements$ ($\env.{coll}\setminus elements$), and the value of all other variables is the same as in $\env$.
More generally, we may consider the \emph{replacement} operator $\omp$, which can remove elements from a collection and add others in their place.
This operation, marked $\omp$, can be defined through combination of $\ominus$ and $\oplus$:
 \begin{equation}
\label{eq:omp}
      \env \omp_{coll} (old\_elements, new\_elements) \doteq \env \ominus_{coll} old\_elements \oplus_{coll} new\_elements.
 \end{equation}
For example $env\omp_{obstacles}(\{(x_1,y_1,r_1)\},\{(x_2,y_2,r_2)\})$ indicates removal of the obstacle $(x_1,y_1,r_1)$ from the obstacle collection of $\env$, and its replacement with $(x_2,y_2,r_2)$.

\subsubsection{Meta-dynamics}
The space of legal mutations represent the meta-control space.
Meta-controls can be applied on meta-states, which are updated according to the ``meta-dynamics:"
\begin{equation}
\label{eq:meta-dynamics}
     \mathfrak{F}(\scenetraj,mut) \doteq \scenetraj',
\end{equation}
where $\scenetraj \doteq (\env,\statetraj,\obstraj)$ is the input simulated scene; $mut\doteq({coll},\omp,inputs)$ is the selected mutation; and $\scenetraj' \doteq (\env',\statetraj',\obstraj')$ is the output simulated scene, where $\env'=env\omp_p inputs$ is the mutated environment, and $\statetraj'$ and $\obs'$ are the new system trajectory and the observation history, respectively, derived from a new simulated system run in the $\env'$, and comply to the agent models, i.e., \linebreak $\dot{\state'}=f(\state',g(\obs'_t))$, $\obs'_t=h(\state', env')$.

\subsection{Falsification as meta-planning for the meta-system}
Together, the meta-state space and meta-control space we defined comprise a meta transition-system, which encapsulates the examined system. We note that, as the system trajectory is returned from a simulator, we do not have direct control over all meta-state variables---we can only actively manipulate the environment and then observe the effect on the system trajectory. In a systems perspective, this means that our meta-system is ``under-actuated.'' Further, since we consider a black-box controller, this means that we do not have an analytical model for the meta-dynamics. We also note that this meta-system is a discrete and Markovian transition system, even when considering a continuous system.

We are interested returning a meta-state that satisfies the condition $\status(\scenetraj)=0$. This condition defines our region of interest within the meta-state space, which contains all the witnesses of task failure. While this condition is easy to test for a given meta-state, it is not trivial to generate a meta-state from this region.
To solve this problem, we suggest to start from some initial meta-state for the meta-system, and then try to lead it into the region of interest, which represents a goal region. Accordingly, we can present the falsification problem as the motion planning problem of finding a goal-reaching trajectory for the meta-system, whose states are simulated scenes.

\paragraph{Problem definition.} Starting from an initial meta-state $\scenetraj_\textit{init}$, find a sequence of meta-controls whose application leads the meta-system into a goal meta-state $\scenetraj_\textit{goal}$, at which $\status(\scenetraj_\textit{goal})=0$.\\

As we see next, this formulation (visualized in \cref{fig:meta-system}) enables us to use efficient motion-planning techniques in order to search for a solution---potentially by evaluating less inputs than through independent sampling.
We clarify, however, that, unlike typical motion planning, in which we are interested in the \emph{path to the goal region} (and often, an optimal one), the solution to our problem is not the path, but the goal state itself, which expresses the falsifying example. We are hence only interested in finding \emph{a} solution as fast as possible.
The path from the initial meta-state to the goal meta-state simply expresses the solution (environment mutation) process, until converging to a falsifying example. 
Further, this motion planning problem does not contain explicit ``obstacles;'' we may consider the invalid meta-states (where the system trajectory is not consistent with the environment) as ``obstacle regions'' to avoid. Nevertheless, by definition, starting from a valid state and applying meta-controls, we cannot end up in an invalid meta-state.

We should also emphasize that this high-level falsification approach is applicable to general black-box systems, with general inputs and outputs, and not just for autonomous systems; though, to keep the discussion grounded, we will continue by considering the case of autonomous systems, where the inputs and outputs are as described---adaptation to other systems is straightforward.
Conveniently, as we explain next, in this case, the incrementality of test-case generation will allow us to share calculations across tested scenes (inputs), resulting also in less control loops (simulation effort) per tested scene, and an even more efficient solution.

\subsection{Exploiting incremental simulation for improved per-test efficiency}
\label{sec:inc-simulation}
\begin{figure}[ht]
    \centering
    \tiny    
    \begin{tabularx}{\textwidth}{|p{0.1\linewidth}|X|X|X|}

    \hline
    & \textbf{Simulation at initial timestamp} & \textbf{Simulation up-until timestamp $T$} & \textbf{Simulation at termination}\\\hline
\raisebox{18pt}{Original scene} 
    &\raisebox{-2pt}{\includegraphics[width=\linewidth]{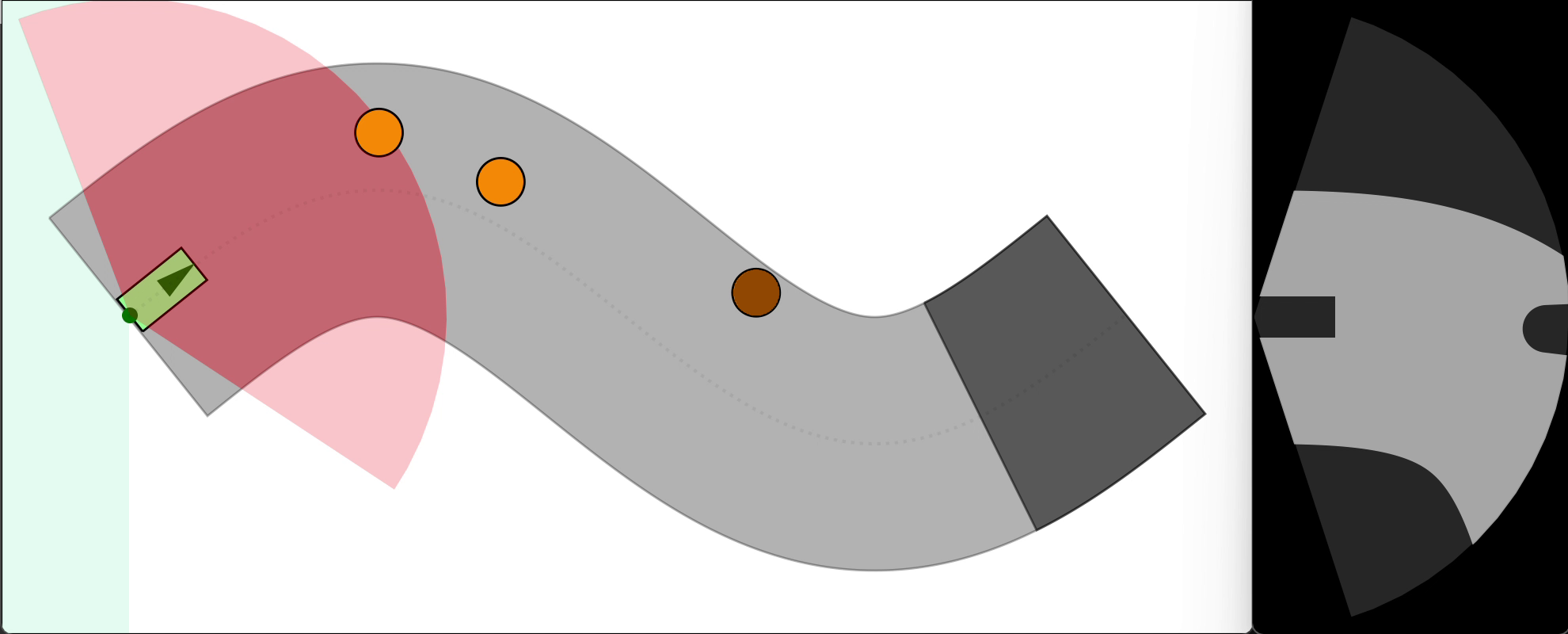}}
    &\raisebox{-2pt}{\includegraphics[width=\linewidth]{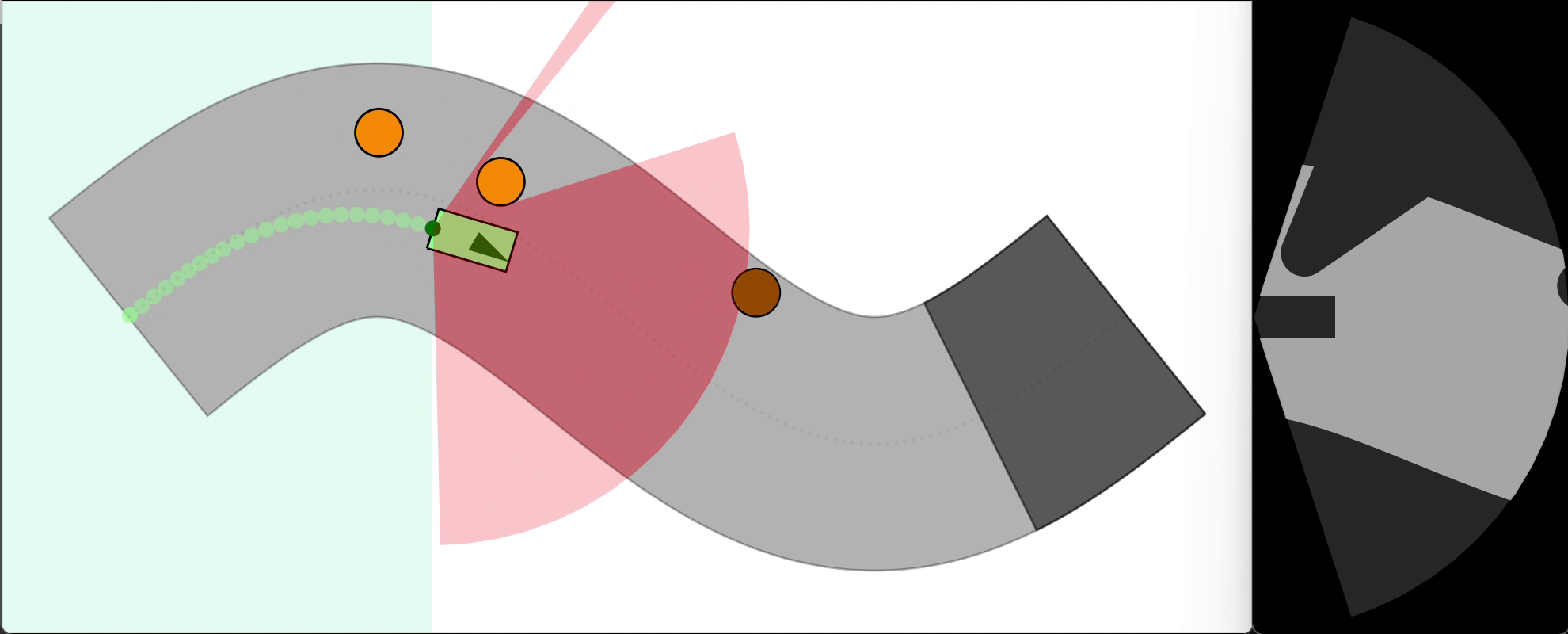}}
    &\raisebox{-2pt}{\includegraphics[width=\linewidth]{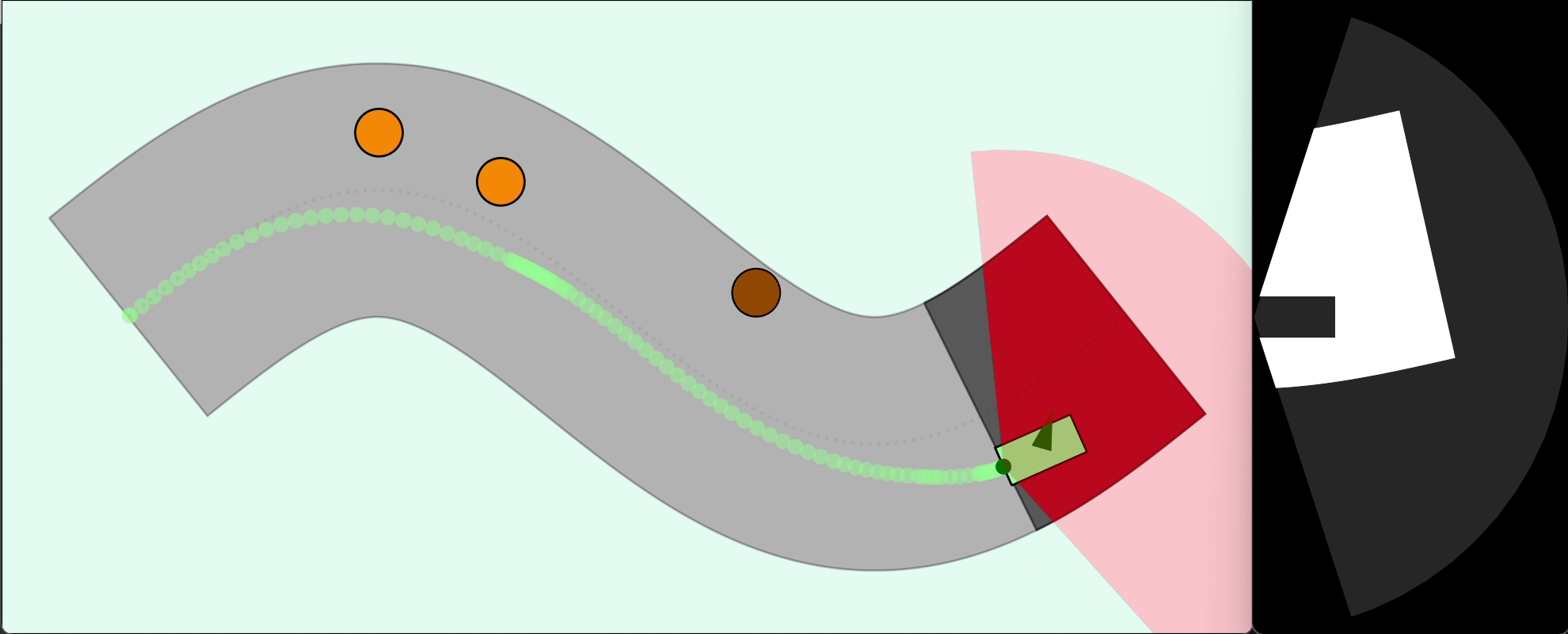}}\\\hline
\raisebox{18pt}{\pbox{\textwidth}{Scene after \\ mutation}}
    &\raisebox{-2pt}{\includegraphics[width=\linewidth]{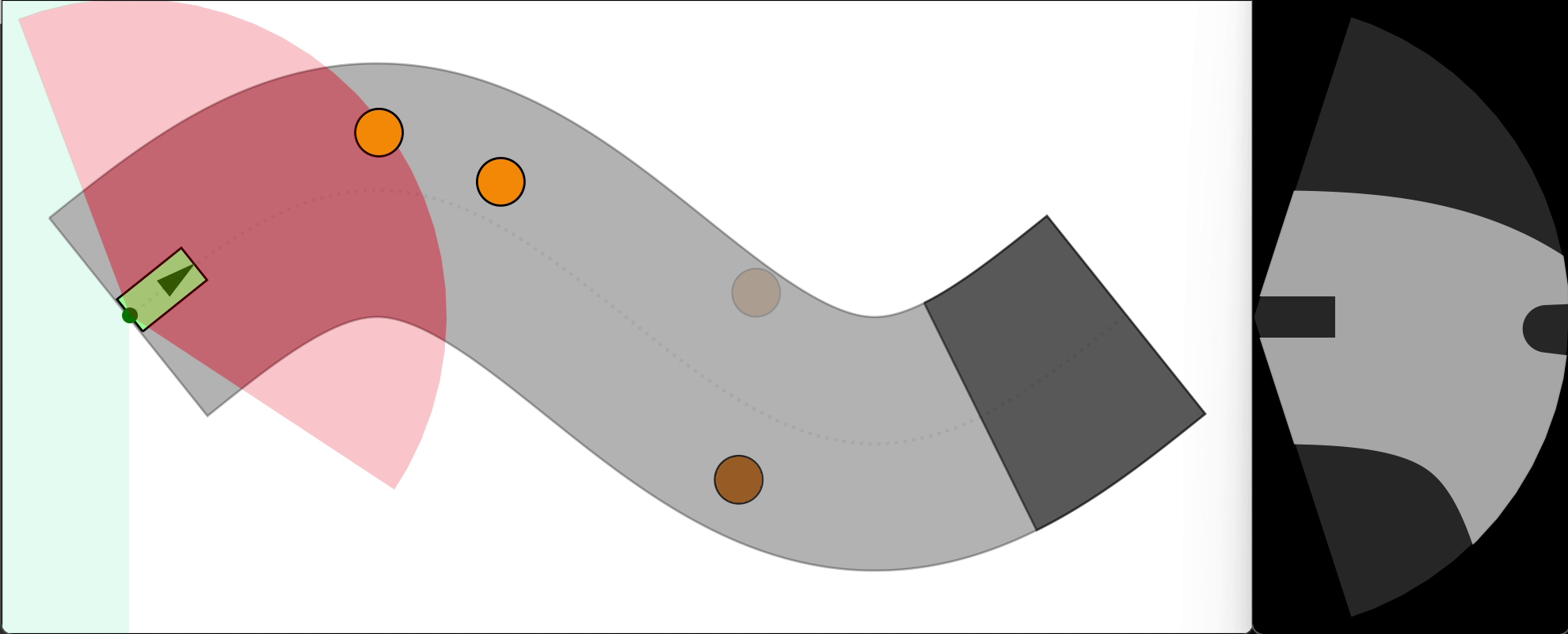}}
    &\raisebox{-2pt}{\includegraphics[width=\linewidth]{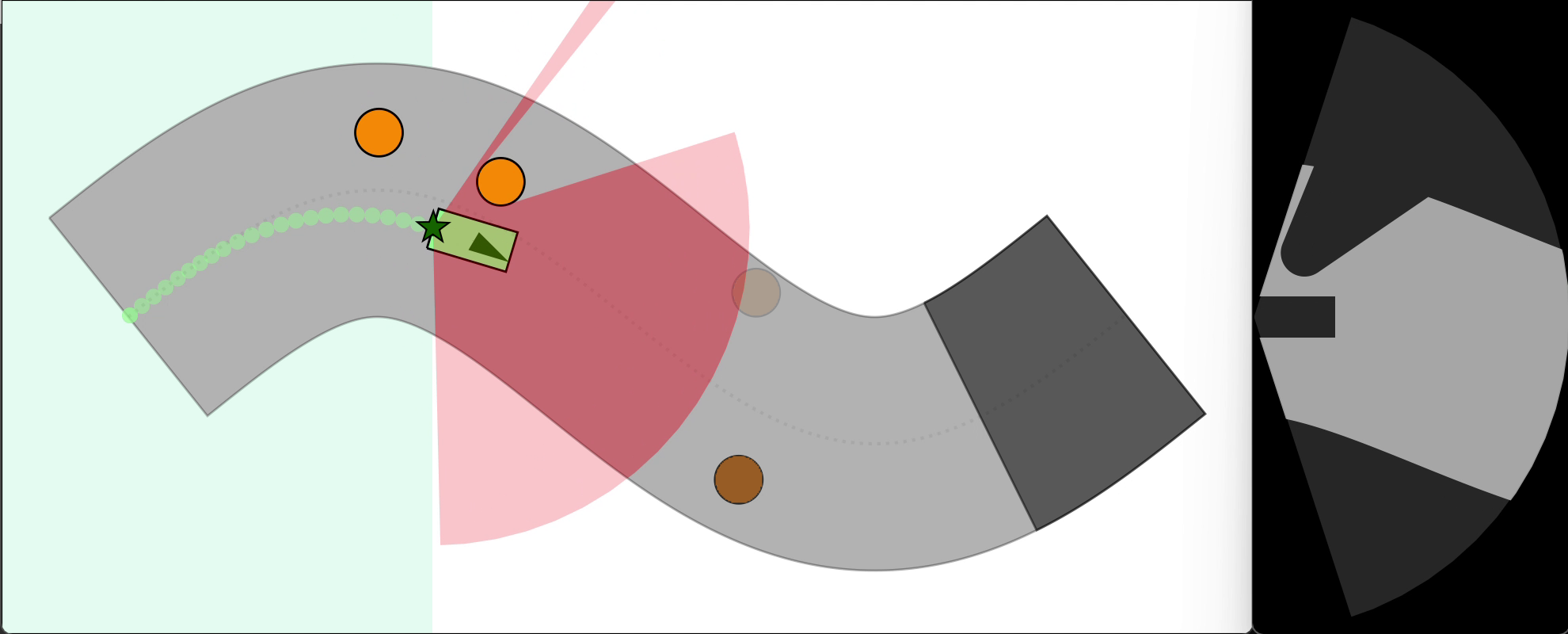}}
    &\raisebox{-2pt}{\includegraphics[width=\linewidth]{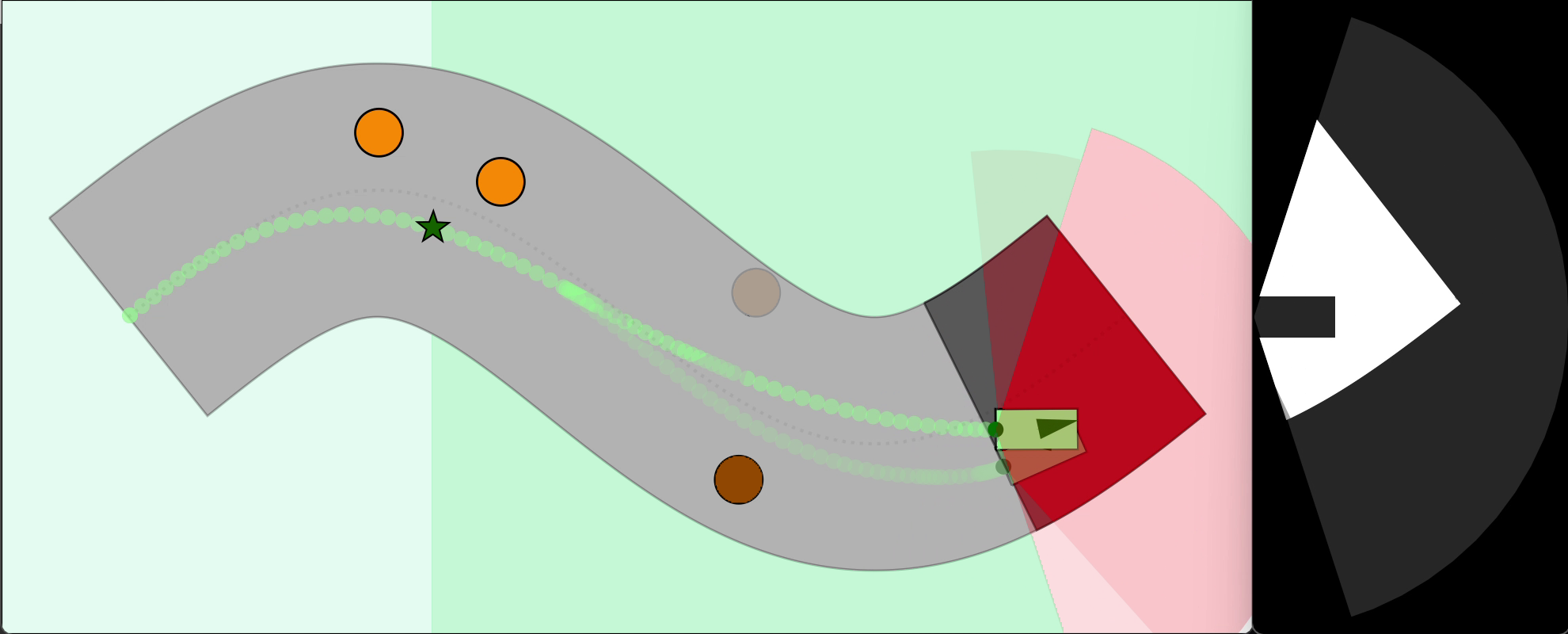}}\\
\hline
\end{tabularx}

    \caption{Comparing the simulation of a scene (environment and initial state), and of the scene after locally mutating it by moving the brown obstacle. The simulation progress is indicated using the green background. The environment mutation only affects the system trajectory starting from timestamp $T$, at which the mutation was observed, and the original observation history was compromised. Thus, to evaluate the mutated scene (calculate the new meta-state), it is enough to run a partial simulation, from car state at timestamp $T$. }
    \label{fig:history}
\end{figure}

As we recall, we would like to minimize the amount of simulation effort needed to find a falsifying goal meta-state.
While our reformulation of the problem as meta-planning should help us reduce the number of tests needed, it can even enable us, when testing autonomous systems, to improve the per-input-test efficiency.
Assuming the environment is only partially-observable by the system, mutating the environment elements may, in general, only partially compromise the validity of the previously-calculated system trajectory---only starting from the timestamp in which this mutation was observed; meaning, system trajectories corresponding to consecutive meta-states share a common prefix.
Hence, in the case of an autonomous system, to avoid redundant calculations when applying meta-controls, after every environment mutation, we may simply incrementally update the previous trajectory, through partial simulation from the identified timestamp (instead of calculating the new trajectory from scratch).
This property, which is illustrated in \cref{fig:history}, allows us to verify the status of each input we reach in our search with less controller queries than it would require to verify inputs sampled independently.
We now see that maintaining the system's observation history as part of the meta-state is essential to understanding the timestamp from which the system trajectory is compromised by the environment change.

We can thus amend in this case the previously-defined meta-dynamics model $\mathfrak{F}$ (\cref{eq:meta-dynamics}) to invoke after an environment mutation only a partial simulation (by rolling back the original simulation to the relevant timestamp), and an \emph{incremental} update of the system and observation trajectories:
\begin{equation}
\label{eq:meta-dynamics-inc}
     \mathfrak{F}(\scenetraj,m) \doteq \scenetraj' = (\env',[\statetraj_{0:T},\statetraj'],[\obstraj_{0:T},\obstraj']),
\end{equation}
where $T$ is the last timestamp in which $h(\state_t,\env)=h(\state_t,\env')$, and 
$\statetraj'$ is a trajectory that starts at $\state_T$ and complies to the system models in the mutated environment $\env'$ (i.e., the result of starting the simulation from a scene $(\state_T,env')$).

We can practically calculate $T$ in a domain-independent way, without making any assumptions on the sensor/controller, by using minimal simulation effort.
To do so, we should sequentially go over the states in the original trajectory $\statetraj$, use the simulator to generate a new sensor observation $\obs'_i$ from each state $\statetraj_i$ in the mutated environment $\env'$, and compare this $\obs'_i$ to the original observation $\obs_i$ taken in the original environment $\env$; if they do not match, it means the history is compromised at this timestamp, otherwise, we should continue checking the rest of the trajectory. If the system is uncertain, we can, more generally, check if it is possible to observe $z_i$ in $\env'$.
Note that this ``prefix-validation'' procedure, does not require simulation of the system dynamics or the (NN) controller, and therefore is much less costly than actually calculating the prefix.
Nevertheless, this procedure can oftentimes be performed without simulation at all, but through simple analysis, given knowledge on the sensor model.
For example, in our example, we can validate each observation $\obs_i$ by first calculating (or conservatively estimating) the ``observed area,'' that is, the observed portion $\subseteq\mathbb{R}^2$ of the track, based on the sensor properties (range and angle), the state $\statetraj_i$ from which the observation was taken, and the environment $\env$; then, simply check if any of the added/removed obstacles overlap with that area, compromising the observation.
The two versions of this validation procedure---the generic and domain-specific ones---are summarized in \cref{alg:expansion} (given in Appendix~\ref{appen:algorithms}).

\draft{, while maintaining \emph{compatibility}.
\begin{definition}
Two environment-states $\env^1,\, \env^2$ of the same environment-type are \emph{compatible} and marked \mbox{$\env_1\sim\env_2$} if for each global property $\prop$ $\env^1[\prop] = \env^2[\prop]$. If also for each local property $\prop$ $\env^1[\prop] \subseteq \env^2[\prop]$, we mark $\env^1 \subseteq \env^2$.
\end{definition}
}

    \section{Solving the problem: sampling-based forward search in the meta-state space}
\label{sec:solving}
\begin{figure}[ht]
    \centering
\includegraphics[scale=0.3]{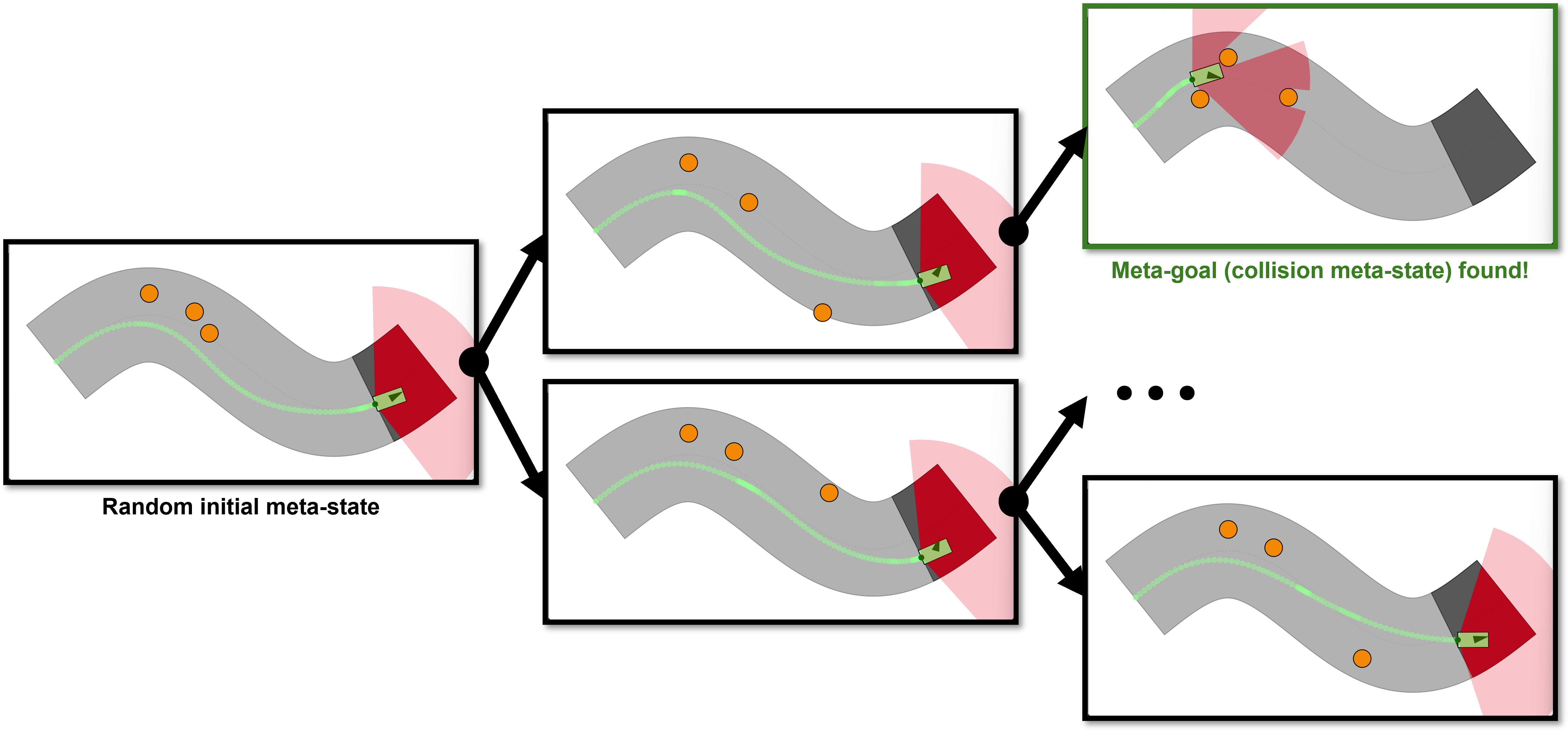}

	\caption{Solving the meta-planning problem: a forward search-tree, where meta-states (nodes) represent simulated scenes, and meta-controls (edges) represent scene mutation followed by incremental re-simulation.}
    \label{fig:forward-search}
\end{figure}

As our innovation is in the reformulation of the falsification as a (meta-)planning problem, our flexible approach is not tied to a specific planner. We can potentially use various off-the-shelf planners to solve the meta-planning problem. 
Since the state space and control space of our meta-system are infinite, to solve the motion planning problem, we suggest to resort to sampling-based motion planning approaches \cite{Orthey2024SamplingBasedMotion,Elbanhawi2014SamplingBasedRobot}.
Further, as we recall, we do not have an explicit model for the meta-dynamics of our meta-system and must rely on a simulator to apply meta-controls. Thus, we do not have access to a ``steering function'' or a ``local planner'' often used by such planners to solve the Two-Point Boundary Value Problem (TPBVP) (i.e., finding local meta-controls connecting two meta-states) \cite{Li2016AsymptoticallyOptimal}. We thus cannot use techniques that rely on backward planning, nor on road-map-based planners. Besides that, road-map-based approach rely on drawing numerous samples from the (meta-)state space, which, in our case, is essentially equivalent to a ``naive" falsification approach, in which we sample and test inputs independently; backward planning relies on having access to a goal-state, which we do not have. Overall, we may only resort to planners that build a forward-search tree in the (meta-)state space, as illustrated in \cref{fig:forward-search}.

Sampling-based forward-search planners \cite{Cortes2020SamplingBasedTree} rely on two operations, which we call alternately for building the search tree: (i)~tree-node selection, and (ii)~tree-node expansion. We will next explain how to perform these operations in our setting, and how we can incorporate basic domain-knowledge to effectively guide and accelerate the search. 
The procedures for node selection and expansion are summarized respectively as \cref{alg:selection} and \cref{alg:expansion}, in 
Appendix~\ref{appen:algorithms}, where we also summarize the overall algorithm for meta-planning-based falsification using sampling-based forward search as \cref{alg:fals}; there, flags for user-enabled options, as we detail ahead, will be highlighted in red.

\paragraph{Meta-planning vs. planning.} To clarify, both our meta-planning approach and the established ``planning'' approaches covered in \cref{sec:related-work} utilize motion planning algorithms for their ability to effectively search a high-dimensional space. Yet, those planning approaches build a planning-tree in the system-state space, where each node represents a system state---until a satisfying branch is found. In contrast, meta-planning builds a planning tree in the composite (input $\times$ output) meta-space, where each tree node (meta-state) represents a scene and a full system-trajectory---until a satisfying node is found. 
In planning, each node expansion requires a short, time-bounded simulation, to generate a trajectory segment; in meta-planning, while we may leverage incremental simulation during node expansion, to avoid recalculating the trajectory prefix, simulations are still performed until termination and results in a full trajectory.

\subsection{Node selection: biasing search towards goal}
\label{sec:goal-bias}
Unguided strategies for section of nodes (meta-states, to be expanded), such as Breadth/Depth-First-Search, or random selection, might take a long time to converge to a solution.
To guide the search towards the goal, and hopefully accelerate it, we can prioritize the selection of nodes for expansion based on their distance-to-goal. In our context, in which the goal of the search is a witness of failure, a distance-to-goal of a meta-state should indicate how ``close'' (we think) the corresponding system was to failing the task, based on its trajectory; we refer to this as the \emph{distance-to-failure} heuristic.
This heuristic can conveniently be encoded by overloading the $\status$ predicate: instead of a binary output, we can allow it to return a value in the range $[0,1]$ conveying the (normalized) heuristic value. It is possible but surely not required to use the robustness function as the heuristic. In general, this heuristic does not need to be smooth or in any restricted format.

In our running example, an intuitive distance-to-failure heuristic can be derived by examining all the states in the car trajectory, finding and returning the car's minimal \emph{distance-to-collision}, i.e., the minimal Euclidean distance of the car shape from an obstacle or the track shoulders along its trajectory. It is possible to choose a different heuristic.

Finally, note that some planners (e.g., RRT \cite{LaValle2001RandomizedKinodynamic}, as we discuss next) sometimes rely on sampling of a goal state and measuring a state-to-state distance from tree nodes to it, in order to bias the node selection towards the goal. This, as we recall, is not possible in our case; thus, even if planning with such a planner, we would still rely on a distance-to-failure heuristic, which conveys the estimated distance to the goal \emph{region}.

\subsection{Node selection: adding exploration}
While beneficial, relying exclusively on a greedy node selection strategy might cause us to get stuck in a dead-end of a (distance-to-failure) local minimum.
So, to ensure complete and efficient solution, it is important to allow for exploration in node selection. As a parameter of the planning algorithm, we can define a ratio between iterations of greedy node selection, and of exploratory selection (according to some strategy), known as the ``goal bias.'' 
The most basic exploration strategy is to use random sampling of tree nodes during exploration iterations. This technique requires no domain-knowledge, though it might not promote a very effective exploration of the search space.
Instead, modern sampling-based motion planners rely on domain-specific information, e.g., a state-to-state distance function and a state-sampling procedure by RRT \cite{LaValle2001RandomizedKinodynamic} and EST \cite{Hsu2002RandomizedKinodynamic}, or a state projection function for density estimation by KPIECE \cite{Sucan2012SamplingBasedTree}, to guide the search and ensure active exploration of the search space. If available, providing such domain-knowledge would allow us to use more advanced, standard planning techniques for an efficient solution.

\subsubsection{Measuring distance between meta-states}
Let us discuss how we can define a meta-state distance function to enable usage of planners like RRT or EST.
In the context of EST, this function shall be used to estimate the density of expanded nodes, in order to prioritize nodes in less-explored regions. In the context of RRT, this function shall be used to measure the distance between nodes in the search tree to a randomly-sampled meta-state; we would then select for expansion the tree-node closest to the sampled meta-state, as a technique for growing the tree toward less-explored regions.

Regardless of our specific problem domain, we can use a composite distance function of the following form between two meta-states $\scenetraj^1, \scenetraj^2$:
\begin{equation}
\label{eq:meta-state-distance}
    \texttt{meta\_state\_distance}(\scenetraj^1, \scenetraj^2) \doteq w\cdot\texttt{env\_distance}(\env^1, \env^2) + (1-w)\cdot\texttt{traj\_distance}(\statetraj^1,\statetraj^2),
\end{equation}
where $w\in[0,1]$, the first component accounts for the distance between the respective environments, and the second---distance between the respective system trajectories.

We recall that our formulation assumes that meta-controls can only modify each environment's elements (e.g., obstacles), which are organized in collections (e.g., sets of obstacles). Thus, we can define the following environment-distance function, which, for a pair $\env^1, \env^2$ of environments of the same type $\envspace$, only accounts for the difference between their respective element collections:
\begin{equation}
    \texttt{env\_distance}(\env^1, \env^2) \doteq
        \sum\limits_{coll\,\in\,\text{element collections of }
        \envspace}\delta_{\textit{coll}}(\env^1.\textit{coll},\env^2.\textit{coll}),
\end{equation}
where for each element collection ${coll}$ defined in $\envspace$, $\delta_{\textit{coll}}$ is a distance function defined between the corresponding collections $\env^1.\textit{coll}$ and $\env^2.\textit{coll}$.

Although we may use a simple, domain-agnostic set-difference function, the function $\delta_{\textit{coll}}$ would benefit from a domain-specific implementation.
In our running example, this function should represent the distance between sets of obstacles on the track, corresponding to $\env^1$ and~$\env^2$.
As an example, we chose a geometric function, aimed at estimating of the overlap between the two sets in the 2D plane. The function is calculated as follows: first, we choose a random point inside a random obstacle from the first obstacle set and measure its Euclidean distance to its nearest obstacle from the second set; we repeat this process multiple times and average the distance, to overall estimate the distance of the first set from the second. Since this distance is not symmetric, we should use a similar process to calculate the distance of the second set from the first, and finally return the average between the two.
It is possible to choose a different distance function.

To complete the meta-state distance function, we require a distance function $\Delta(\statetraj^1,\statetraj^2)$ between system trajectories. In the context of our running example, since the trajectories can, generally, be of different length, we cannot simply rely on measuring distance between corresponding states. Yet, since the system motion is continuous and the states are geometric, we may use, for example, a function based on distance-between-curves. We chose to use the 2D area bounded between the curves, calculated by estimating the integral of the difference between the curves (parameterized as a function $y=f(x)$):
\begin{equation}
\label{eq:traj-distance-ex}
\texttt{traj\_distance}(\statetraj^1,\statetraj^2) \doteq \int_{x_\text{init}}^{x_\text{final}} \left[ \max(\statetraj^1(x),\statetraj^2(x))-\min(\statetraj^1(x),\statetraj^2(x)) \right] dx.
\end{equation}

\subsubsection{Sampling of meta-states for planning with RRT}
In the general case, sampling a meta-state is done by sampling a scene and running a simulation of the system. In our running example, where the initial system state and environment parameters are set, sampling a scene simply means sampling locations for the obstacles on the track.

We note that, for RRT-based planning, we cannot directly sample non-goal meta-states, as required by the algorithm for exploratory iterations.
Thus, for such iterations, we should sample a random meta-state, and then check whether it is a goal meta-state: if it is not (the likely case), then we would simply expand the tree towards this meta-state; if it is a goal state, then we actually \emph{found a solution} to our problem, and we can quit the search.
For the same reason, as already mentioned, in the goal-biased iterations, we rely on the distance-to-failure heuristic for node selection. Meaning, somewhat unconventionally, we shall use two different distance functions: one to grow towards the goal region, and second to grow towards random samples during exploration extensions.

\subsection{Node selection: simplified}
\subsubsection{Simplified distance between meta-states}

We recognize that measuring system-trajectory distance (in order to calculate the meta-state distance (\cref{eq:meta-state-distance}) may not be trivial, in terms of both the definition of a proper distance, and the effort required to calculate it (as in \cref{eq:traj-distance-ex}).
We can, thus, consider a simplified meta-state distance function, which ignores the system-trajectory distance, and only accounts for the environment distance. Such a simplification means that our meta-state distance function measures distances in a projection of the meta-state space, which abstracts away the system trajectory (simulation output). In practice, we can do so by returning an uninformative zero distance output for every pair of system trajectories.
\draft{Another, less-severe option, if we do have a state-to-state distance function, is to only measure the distance between the final states in the respective agent trajectories; in our running example, we can, for example, measure the difference in progress along the track (a scalar).}

\subsubsection{Simplified sampling of meta-states for planning with RRT}
We also recognize that meta-state sampling, as described before for planning with RRT-based node selection, involves sampling of scenes and simulating them, and hence incurs a cost in terms of simulation effort. Further, since these meta-state samples are independent of each other, we cannot benefit from the aforementioned incremental computation, making this cost potentially significant. 
Nevertheless, as the simplified distance function described above ignores the simulation output anyways, when using it for node selection, there is also no need to waste computational effort on simulation of the sampled scenes.
Meaning, in that case, we should coincidently rely on a simplified meta-state sampling procedure, which returns non-simulated scenes (in our case, by just sampling the track's obstacle collection).
Since theses scenes can be viewed as redundant meta-states with a trajectory of length one, they are still in the meta-state space, and this choice can simply be presented as domain-agnostic biasing of the sampling procedure.
With this simplification, simulation effort is only invested in node expansion, and a solution can only be discovered during that phase of the tree growth.

\subsection{Node expansion}
Consider the general case in which our meta-control space is based on ``replacement'' ($\omp$) mutation operations (\cref{eq:omp}), i.e., in our running example, removing existing obstacles and adding new ones in their place. 
As mentioned, in our meta-system, we do not have access to a steering function; meaning, we cannot easily find meta-controls to take our meta-system to an exact destination (e.g., the goal region or a sampled meta-state). Taking notes from the ``kinodynamic planning" literature \cite{LaValle2001RandomizedKinodynamic,Li2016AsymptoticallyOptimal,Sucan2012SamplingBasedTree,Elbanhawi2014SamplingBasedRobot}, in cases where we only have access to a ``forward propagation'' model, we may simply consider extension of a randomly selected meta-control(s) from the selected node during node expansion. This assumption should not sacrifice the probabilistic completeness of the algorithm \cite{LaValle2001RandomizedKinodynamic}.

As further recognized by the ``kinodynamic planning" literature \cite{LaValle2001RandomizedKinodynamic,Sucan2012SamplingBasedTree}, the step-size for the sampled controls may either be random or predetermined. Yet, unlike kinodynamic planning, where the step-size is determined by a simple scalar expressing time duration (there,~$\Delta t$), in our formulation, the step-size corresponds to the more-complex ``size of environment-mutation;'' we identify two user-controllable parameters, in two axes, that govern this size.
The specification of these parameters, to control the step-size, can practically be encoded as a ``biased'' action sampler.

\subsubsection{Environment-mutation width}
The first parameter accounts for the number of elements (in the case of our running example---obstacles) we replace in the environment. The user may choose to consider a constant number (e.g., every meta-control always replaces only a single, randomly-chosen obstacle), or randomly choose one (e.g., replace a randomly-selected subset of a random size of obstacles, according to a uniform distribution); for brevity, we refer to the first option as ``constant width'' and to the latter as ``random width.'' Choosing the latter option makes it possible to occasionally sample a meta-control that replaces all the  environment elements in the collection (obstacles).
With that, we can essentially restart a new search tree from a newly-sampled initial meta-state. This can be effective for improving exploration and help quickly pulling the search tree away from local minima.
We may choose to allow such steps only in ``exploration iterations'' or also when trying to steer towards goal. \draft{Can I not just use this to ensure exploration and always run a greedy tree?! no need for RRT or random node selection at all... checked this - it was less effective}
Notably, unlike kinodynamic planning, the meta-control sampling is \emph{dependent} on the selected step-size: we should first determine how many obstacles to replace (i.e., the mutation width), and then how to replace them.

\subsubsection{Environment-mutation depth: environment element perturbation}
To sample a random meta-control, after choosing the mutation-width $n$, we shall randomly select a subset of $n$ elements (in our example, obstacles) to replace in the environment, and finally randomly select their replacement.
The ``distance'' between the elements (obstacles) to be replaced, and their corresponding replacements, conveys the ``mutation depth.''
Currently, we may assume the selection of the subset of elements for replacement is uniformly random.
The simplest way to then determine this replacement, with no need for specialized knowledge, is to uniformly sample a new environment element for each element removed. Yet, allowing each removed element to be replaced with a completely arbitrary new element would mean that the mutation depth can be arbitrarily large. This, in turn, would mean that even if the mutation width is low, the distance between the initial meta-state and the posterior one, after applying the mutation, can also be arbitrarily large, and at an uncontrollable direction. This can hinder our ability to incrementally guide the search-tree growth.
A better alternative would be to locally perturb each element chosen for removal---yet, this is dependent on the availablity of such a perturbation procedure. In our example, perturbation of the geometric location of chosen obstacles, e.g., by adding Gaussian white noise to their original location, is easy and trivial; the standard deviation of this Gaussian noise distribution would control the tightness of the perturbation and, by such, the potential mutation depth.

While this will not be evaluated in this paper, given additional, ``gray-box'' domain-knowledge, we may further inform and bias the meta-control sampling procedure, e.g., based on the meta-state distance function calculation, by actively identifying environment mutations that are most likely to challenge the system, or by identifying weak points in the system trajectory.

\subsection{On the usage of domain knowledge}
In the previous sub-sections, when discussing the planning procedure, we covered several planner parameters the user can specify in order to control the and direct the search-tree growth.
It is important to differentiate between domain-agnostic planner parameters, and domain-specific knowledge/procedures.
Specifically, the specification of the mutation-width sampling distribution, and choice of meta-control sampling technique (uninformed replacement vs. perturbation) during node expansion; the choice of node selection technique (e.g. random, greedy, or RRT); and the goal-bias ratio and choice whether or not to rely on simplified distance function for (RRT-based) node selection are all \emph{not} considered domain-knowledge.
Still, to practically enable some of these options, we do require domain-specific procedures, including an environment-element perturbation procedure, distance-to-failure heuristic, environment distance, and system-trajectory distance. 

Our formulation allows us to gradually integrate such domain knowledge to improve the search, based on its availability (as we demonstrate in the experimental results to follow), and can even work with no domain knowledge at all.
This comes in contrast to standard techniques covered in \cref{sec:related-work}, which \emph{require} the user to non-trivially define the robustness function, and/or provide hard-to-achieve probabilistic knowledge on prior distributions, and/or have access to the system model.
The type of domain knowledge we might consider is mostly heuristic and comparative, which is arguably more intuitive, less-specialized, and easier to obtain.
For example, the distance-to-failure heuristic function we consider only retrospectively scores a given scenario, to indicate if one system run seems ``closer'' to a failure than another---it does not require knowledge on how to generate a failure scenario; our environment distance function only indicates how ``similar'' one environment is to another---it does not require knowledge on the distribution of adversarial environments in the environments space; our meta-control sampler only requires knowledge on how to locally modify environment elements---it does not not require an understanding of which elements should challenge the system.

Overall, we do not assume any prior knowledge on the transition dynamics in the meta-space, the system's dynamic model in the environment, nor of the way the system interacts with the environment. We also, as we recall, make no assumptions on the NN controller nor on the way it was trained.
    
    \section{Experimental evaluation}
\label{sec:evaluation}
We tested our falsification approach (summarized in \cref{alg:fals}) in the context of our running-example system described in \cref{sec:running-example}: an autonomous car on an obstructed track environment, trying to navigate to its end zone while avoiding collision. 

\subsection{Scenario}
To falsify the autonomous car, we were searching for a placement $(x,y)\in\mathbb{R}^2$ for three circular obstacles of radius $r=0.1$ on a sinuous track, which would cause the car to steer into collision.
Simulations of this scenario were conducted using our lightweight and open-source Python engine LiteRacer \cite{Elimelech2024LiteRacerLightweight}. A screenshot from one of the simulations is provided in \cref{fig:car}.

The track parameters are assumed to be set: the track shape is the padded area around the curve $y=0.8\cdot \sin(x)$ in the range $x\in[0,5\pi]$ units, with track width set to $1.6$ units. The end zone is defined as the area in the range $x\in[4.5\pi,5\pi]$.
The other scenario properties were set as follows:
the car bounding box is of size $0.2\times0.4$ units, with its frame of origin located in between the back wheels. The initial car position is at $(x=0,y=0)$, with the heading set to match the track curve direction, and the steering angle set to $0$. The car maximum speed is capped at $0.4$ units/second, while the steering speed is limited to $[-10,10]$ degrees/second, and the steering angle is limited to $[-60,60]$ degrees. The control frequency is 1Hz. The sensor is placed between the front wheels; the sensor range is $2$~units, the sensor angle is in the range $[-72,72]$ degrees, and the observation image resolution is $100\times50$ pixels.

The controller was originally trained through repeated randomly-seeded simulations of the same environment using AI Gym \cite{Brockman2016OpenAIGym} and the StableBaselines~\cite{Raffin2021StableBaselines3Reliable} implementation of the Soft Actor-Critic (SAC) algorithm, until apparent convergence to the desired behavior. We, nonetheless, treat this controller as a black-box.

\subsection{Comparison}
To study and prove the benefit of our approach, we compared the computational effort required by different algorithms, both baselines and variations of our algorithm, to falsify the controller in the described scenario.
For each algorithm, we ran 20 randomly-seeded experiments and measured the effort invested in each one.
 We present the effort in terms of both the number of controller calls, which indicates the simulation duration, and the number of environments examined until finding a falsifying one. For the baseline approaches, the latter measure indicates the number of independent environment-samples until finding a falsifying one; for meta-planning, this measure indicates the number of nodes in the search tree. Though, it is important to consider that in meta-planning, thanks to the incrementality of the simulation, each environment in the planning tree is only \emph{partially} examined. Each algorithm assumed access to different amounts of domain knowledge, as indicated later in \cref{tbl:results}, while all have access to a basic environment (input) sampling procedure.

Overall, we compared eight different falsification algorithms: the first, with uniform, uninformed environment sampling, serving as a basic baseline; the second, using a genetic algorithm (as prescribed in \cite{Zhao2003GeneratingTest}); the third, using Bayesian Optimization (as prescribed in \cite{Deshmukh2017TestingCyberPhysical}); and the remaining five are variations of our meta-planning-based falsification algorithm---gradually increasing the amount of domain-knowledge available to the algorithm.
The variations of our approach are detailed as follows:
\begin{enumerate}
\item Unguided meta-planning with no limitation of the allowed environment mutation (meta-control) depth; this approach assumes no domain knowledge, like the uninformed baseline.
\item Unguided meta-planning with limited mutation depth; this requires a procedure to locally perturb an environment.
\item Goal-biased greedy meta-planning; this also requires access to a distance-to-failure heuristic.
\item Goal-biased RRT-based meta-planning with simplified exploration and sampling procedures; this also requires access to environment-sampling and environment-distance procedures.
\item Unsimplified goal-biased RRT-based meta-planning; this also requires access to a trajectory distance procedure.
\end{enumerate}

For all these variations, we extended a single random meta-control with an unlimited mutation width during node expansion, corresponding to replacement of a random subset of the obstacles collection. For the first variation, the mutation depth was unlimited, meaning, the subset of obstacles could be replaced with uniformly-sampled new obstacles.
For all other variation, the mutation depth was limited to a local perturbation, defined by adding white noise (with standard deviation$=(2,2)$) around the original location of each replaced obstacle.
For the unsimplified RRT planner, we measured how many controller calls came from meta-state sampling for node selection (only required for this variation), and how many came from applying meta-controls during node expansion (as done in all other variations).

When implementing the genetic algorithm, we used a population of $4$ environments in each generation. To evolve the population, we used both crossover operations between environment pairs (random merger of the obstacle sets of two environments), and perturbation-mutation operations for individual environments (as we use for meta-control). In each new generations, two of the samples were generated through crossover, and two through perturbation. In both cases, the selection of environments for mutation relied on a ``fitness function'' based on the distance-to-failure (as suggested in \cite{Zhao2003GeneratingTest})---effectively acting as a goal bias. Of course, implementation of these operations required some domain knowledge.

To implement the Bayesian-Optimization-based falsification algorithm, we used the open-source ``Bayesian Optimization'' Python package \cite{Nogueira2014BayesianOptimization}.
As explained, BO (and other optimization-based falsifiers) cannot natively be used in our described scenario, and required several adaptations. First, we used our testing model, in which the input is the scene and the observation signal is generated during the simulation.
Second, as mentioned in \cref{sec:running-example}, in this scenario the robustness function over the car trajectory is not well-defined; thus, we used our distance-to-collision heuristic as the optimization objective---a sensible replacement considering these two functions are meant to measure similar quantities. 
Third, as the number of variables must be predefined, we modified the environment to include three explicit vectors to encode the obstacle locations, instead of using a flexible collection. Fourth, the approach requires defining many other parameters, such as kernels and an acquisition function; for that, we used the default options suggested by the BO package. Fifth, we needed to consider an alternative and less-trivial parametrization for the obstacle location, as the $(x,y)$-parameterization does not correspond to a continuous box; we incorporated expert knowledge to define the ``distance along the track'' and the ``closeness to the right shoulder'' as the alternative parameters. 
Our implementation of all falsification algorithms is available at \cite{Elimelech2024LiteRacerLightweight}.

\begin{table}[ht]
\caption{Summary of results: average computational effort taken to find a falsifying environment using different algorithms. Lower is better. The numbers in brackets in the bottom row represent the effort for tree expansion only, not including the sampling effort.}
\label{tbl:results}
\centering
\tiny
\begin{tabular}{|p{1.4cm}!{\vrule width 1pt}p{1.1cm}!{\vrule width 1pt}p{2.5cm}!{\vrule width 1pt}p{1.9cm}p{1.4cm}!{\vrule width 2pt}rrrr|}
\hline
\multirow{3}{*}{\textbf{\pbox{\textwidth}{Algorithm \\category}}} & \multirow{3}{*}{\textbf{Algorithm}} & \multirow{3}{*}{\textbf{Domain knowledge}}                                                                              & \multicolumn{2}{c!{\vrule width 2pt}}{\textbf{Algorithm options}}                                                                                     & \multicolumn{4}{c|}{\textbf{\pbox{\textwidth}{Average effort to find \\a falsifying environment}}}                                                                                                                                                                                                                                          \\ \cline{4-9} 
                                             &                                     &                                                                                                                         & \multicolumn{1}{l|}{\multirow{2}{*}{\textbf{\pbox{\textwidth}{Meta-control /\\ env. mutation type}}}} & \multirow{2}{*}{\textbf{Goal bias}}               & \multicolumn{2}{c!{\vrule width 1pt}}{\textbf{Envs. tested}}                                                                                                                     & \multicolumn{2}{c|}{\textbf{Control loops}}                                                                                                    \\ \cline{6-9} 
                                             &                                     &                                                                                                                         & \multicolumn{1}{l|}{}                                                           &                                                   & \multicolumn{1}{r|}{\#}                                                            & \multicolumn{1}{r!{\vrule width 1pt}}{\%}                                                          & \multicolumn{1}{r|}{\#}                                                          & \%                                                          \\ \noalign{\hrule height 2pt}
Random search                                & Uniform sampling         & None                                                                                                                    & \multicolumn{1}{l|}{---}                                                          & No                                                & \multicolumn{1}{r|}{182.05}                                                        & \multicolumn{1}{r!{\vrule width 1pt}}{100\%}                                                       & \multicolumn{1}{r|}{11564}                                                       & 100\%                                                       \\ \noalign{\hrule height 2pt}
\pbox{\textwidth}{Optimization \\(passive)}                       & \pbox{\textwidth}{Genetic \\algorithm}                   & \pbox{\textwidth}{Value function \\+ env. perturbation procedure\\    + env. crossover procedure} & \multicolumn{1}{l|}{\pbox{\textwidth}{Env. perturbation, \\ two-env. crossover}}                    & Yes (via fitness function)                        & \multicolumn{1}{r|}{193.80}                                                        & \multicolumn{1}{r!{\vrule width 1pt}}{106\%}                                                       & \multicolumn{1}{r|}{10030}                                                       & 86\%                                                        \\ \noalign{\hrule height 2pt}
\pbox{\textwidth}{Optimization \\(active)}                        & \pbox{\textwidth}{Bayesian \\optimization}               & \pbox{\textwidth}{Value function \\+ assumptions (listed in text)}                                                                          & \multicolumn{1}{l|}{---}                                                          & Yes (via learned model)                           & \multicolumn{1}{r|}{123.14}                                                        & \multicolumn{1}{r!{\vrule width 1pt}}{68\%}                                                        & \multicolumn{1}{r|}{7901}                                                        & 68\%                                                        \\ \noalign{\hrule height 2pt}
\multirow{5}{*}{\pbox{\textwidth}{Meta-planning \\ (ours)}}        & Random tree                         & None                                                                                                                    & \multicolumn{1}{l|}{\pbox{\textwidth}{Unlimited depth \\(uniform replacement)}}                              & No                                                & \multicolumn{1}{r|}{135.00}                                                        & \multicolumn{1}{r!{\vrule width 1pt}}{74\%}                                                        & \multicolumn{1}{r|}{7192}                                                        & 62\%                                                        \\ \cline{2-9} 
                                             & Random tree                         & Env. perturbation procedure                                                                                      & \multicolumn{1}{l|}{\pbox{\textwidth}{Limited depth \\(perturbation)}}                               & No                                                & \multicolumn{1}{r|}{133.60}                                                        & \multicolumn{1}{r!{\vrule width 1pt}}{73\%}                                                        & \multicolumn{1}{r|}{6537}                                                        & 56\%                                                        \\ \cline{2-9} 
                                             & Greedy tree                         & \pbox{\textwidth}{'' \\+ distance-to-failure heuristic}                                                                                      & \multicolumn{1}{l|}{''}                                                         & Yes (via node selection): 100\%                   & \multicolumn{1}{r|}{115.7}                                                         & \multicolumn{1}{r!{\vrule width 1pt}}{63\%}                                                        & \multicolumn{1}{r|}{5503}                                                        & 47\%                                                        \\ \cline{2-9} 
                                             & Simplified RRT                      & \pbox{\textwidth}{'' \\+ env. distance}                                                                                               & \multicolumn{1}{l|}{''}                                                         & Yes (via node selection): 80\% + 20\% exploration & \multicolumn{1}{r|}{96.05}                                                         & \multicolumn{1}{r!{\vrule width 1pt}}{52\%}                                                        & \multicolumn{1}{r|}{4703}                                                        & 40\%                                                        \\ \cline{2-9} 
                                             & RRT                                 & \pbox{\textwidth}{'' \\+ system-trajectory distance}                                                                                              & \multicolumn{1}{l|}{''}                                                         & ''                                                & \multicolumn{1}{r|}{\begin{tabular}[c]{@{}r@{}}88.75\\ (74.75)\end{tabular}} & \multicolumn{1}{r!{\vrule width 1pt}}{\begin{tabular}[c]{@{}r@{}}48\%\\ (41\%)\end{tabular}} & \multicolumn{1}{r|}{\begin{tabular}[c]{@{}r@{}}4524\\ (3642)\end{tabular}} & \begin{tabular}[c]{@{}r@{}}39\%\\ (31\%)\end{tabular} \\ \hline
\end{tabular}

\end{table}

\subsection{Results}
\begin{figure}[h]
\centering
\includegraphics[width=0.48\textwidth]{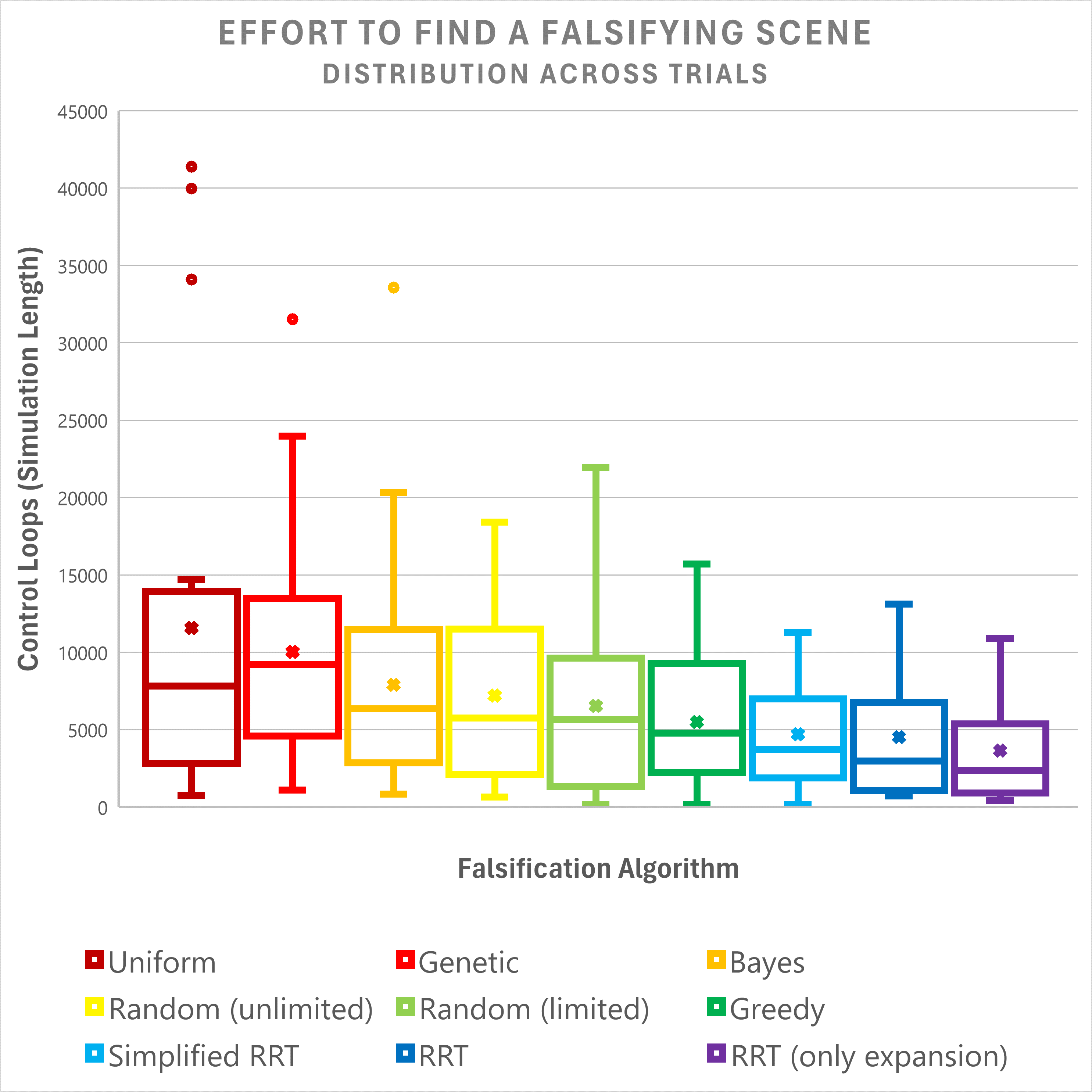}
\includegraphics[width=0.48\textwidth]{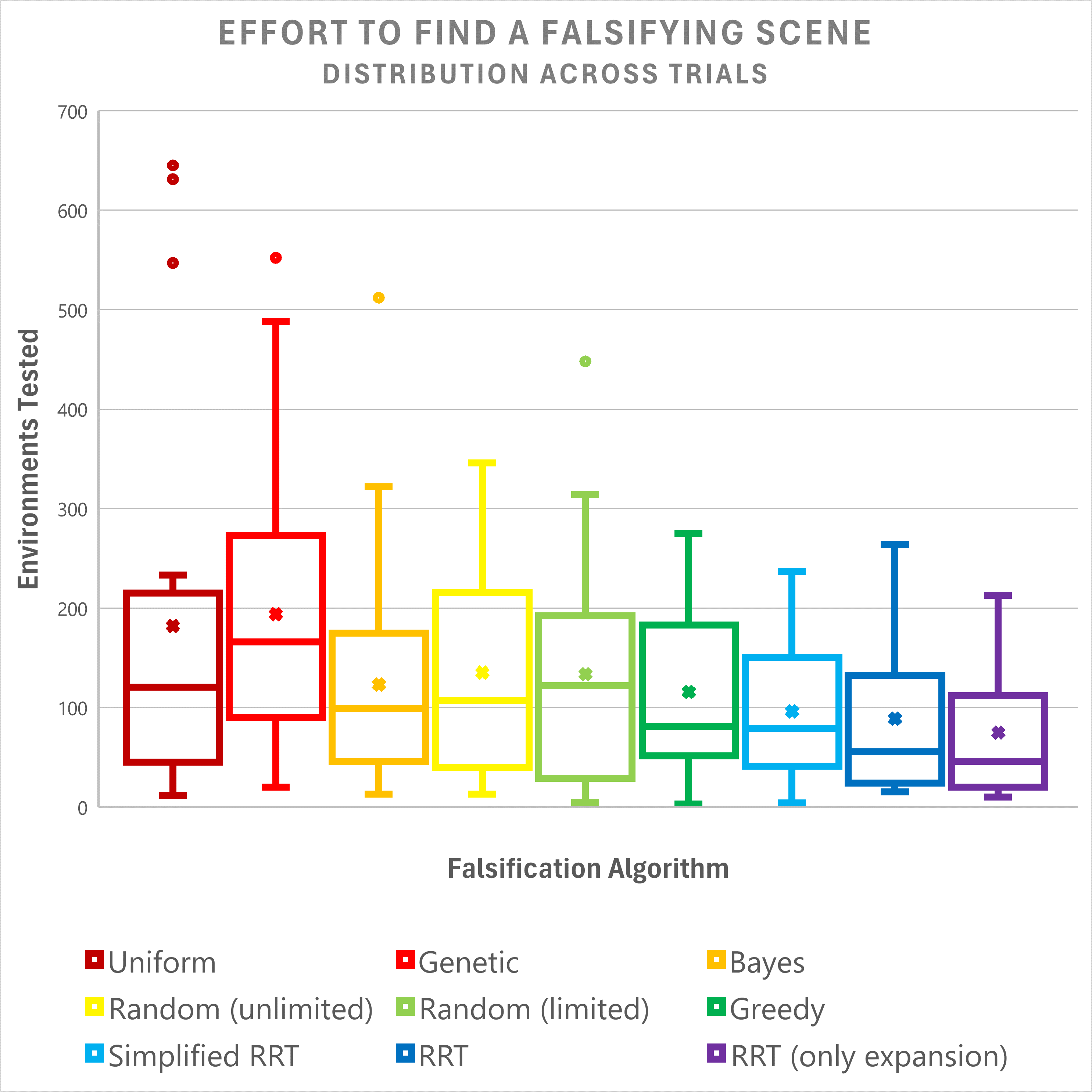}

\caption{``Box and whisker charts'' representing the distribution of effort across trials. Lower is better. Each box represent the two central quantiles of the distribution; the central line represents the median, and ``x'' the mean; the whiskers extend to the maximum and minimum values, excluding ``outliers,'' which are depicted explicitly. Variations of meta-planning (right six columns) beat the baselines (left three) in both metrics.}
\label{fig:results-dist}
\end{figure}

\cref{tbl:results} and \cref{fig:results-dist} showcase the computational cost consumed by each algorithm to find a falsifying environment in an average experiment (lower is better).
These indicate clearly that our approach significantly outperforms the three baselines, both in terms of number of environments tested, and the simulation effort. Even with our simplest variation, which does not require any domain knowledge, we were able to save about 40\% of the computational cost compared to the uniform-sampling-baseline, about 30\% compared to the genetic-algorithm-baseline, and about 10\% compared to the BO-baseline---thanks simply to the incremental nature of our search, and despite the fact that the latter two require more domain-knowledge and are guided by a domain-specific heuristic. With the remaining variations, we can see that incorporation of the domain-specific procedures into meta-planning allowed us to further guide the search and reduce the computational cost.
All of our goal-biased variations beat the BO and genetic-algorithm baselines, both in terms of the number of controller calls and the number of environments examined, despite relying on the same heuristic. Meaning, our meta-planning approach was able to make better usage of the available domain knowledge to guide the search.
Overall, our top variation achieved about 60\% reduction in computational cost compared to uniform sampling, about 55\% compared to the genetic-algorithm, and about 45\% reduction compared to BO.

We can see that although the BO-baseline seems to examine less environments than our non-goal-biased variations, they still outperform it in terms of control loops, thanks to the incrementality of the environment generation (and simulation). More generally, we can see that, unlike the BO-baseline, the reduction in controller calls and in environments examined, in comparison to the uniformed-baseline, is not consistent. Meaning, we see a more significant reduction in the number of control calls, again, since our approach requires less controller calls to evaluate each environment. We can also see that the random tree with depth-limited mutation performed better in terms of control-loops from its unlimited counterpart, despite the two performing similarly in terms of environments; this is because limiting the mutation made subsequent meta-state in the tree to be ``closer together,'' leading to better exploitation of the simulation incrementality. 

With the final variation (RRT), we recall that some of the cost includes also explicit meta-state sampling, in order to guide the node selection process. 
If we only measure the cost invested in tree expansion (numbers in brackets, without counting that sampling), this variation seems to outperform the ``simplified RRT'' variation. This indicates the potential of incorporating the information on both the input and output distributions to better guide the search (less environments examined) in comparison to the simplified variation, which only used information on the input distribution.
Nevertheless, when also counting the meta-state sampling cost (which is not required for the simplified variation), the overall cost becomes similar to simplified variation, as the improvement in search-guidance is balanced the by the additional cost required for node selection. Overall, in this scenario, unsimplified RRT only slightly outperformed its simplified counterpart. 

An extended theoretical discussion on the algorithm and these results is given in Appendix~\ref{appen:theory}.

    \section{Conclusion}
This paper investigated the problem of efficiently finding falsifying inputs to an autonomous system guided by a black-box controller under general specifications.
We first identified that, for such systems that observe their environments using a high-dimensional sensor, the input should correspond to a description of an environment, used to initiate a simulation, rather than a sensor signal directly, as most often considered by existing approaches. We accordingly formulated a more appropriate system model, which inputted the static environment description, and outputted the system trajectory and observation history.
Considering this model, we suggested a novel reformulation of the falsification problem as planning a trajectory for a meta-system, or, in short, meta-planning. Meta-states of this meta-system encapsulate input-output pairs, and transitioning between them is done by applying meta-controls---mutations to the input, which invoke an update to the output.
As we showed, such an update could be performed incrementally in each expansion iteration of the planning algorithm---only performing a partial simulation, starting from the time-step in which observation-history was compromised. 
The meta-planning problem can be solved with off-the-shelf planners, and we specifically provided guidelines on how to employ the sampling-based RRT algorithm for that task.
We finished with experimental evaluation of the approach for the problem of an autonomous car with a NN-controller, trained to lead it to the end zone of a track while avoiding collision with randomly-scattered obstacles.
Our experiments proved that we could significantly reduce the falsification effort, in terms of both the average number of samples, and the average simulation effort (number of control loops) per sample---even over approaches informed by more extensive domain knowledge.

While here we used RRT as our planner, this reformulation opens the door to the development of new specialized planners for this problem, e.g., a planner that can utilize parallelizable simulation.
As we alluded to throughout the paper, in future work, we would also like to extend this approach to support systems under additional sources of uncertainty, e.g., considering also action or observation noise.
Finally, in a very promising future direction, we will seek to leverage properties of this planning-based reformulation to derive (probabilistic) guarantees for state coverage and formal failure estimates.
    
\bibliographystyle{splncs03_unsrt}
\bibliography{Falsification}

\newpage
\appendix
    \section{Algorithms}
\label{appen:algorithms}

\begin{algorithm}[ht]
\scriptsize
\SetAlgoLined\DontPrintSemicolon
\setstretch{1.1}

\SetKwProg{alg}{Algorithm}{}{}
\SetKwProg{proc}{Procedure}{}{}

\SetKwFunction{falsify}{falsification\_as\_meta\_planning}

\SetKwFunction{selectNode}{select\_node}
\SetKwFunction{expandNode}{expand\_node}
\SetKwFunction{dtf}{distance\_to\_failure}
\SetKwFunction{perturbEnv}{perturb\_env\_elements}
\SetKwFunction{replaceEnv}{replace\_env\_elements}
\SetKwFunction{sampleStartScene}{sample\_start\_scene}
\SetKwFunction{sampleEndScene}{sample\_start\_scene}
\SetKwFunction{sampleMetaState}{sample\_meta\_state}
\SetKwFunction{envDist}{env\_distance}
\SetKwFunction{trajDist}{agent\_traj\_distance}
\SetKwFunction{sceneTrajDist}{scene\_traj\_distance}
\SetKwFunction{metaAction}{random\_meta\_control}
\SetKwFunction{simulate}{simulate\_system}
\SetKwFunction{metastatedist}{meta\_state\_distance}
\SetKwFunction{envdist}{env\_distance}
\SetKwFunction{sampleenv}{sample\_env}

\SetKw{True}{True}
\SetKw{False}{False}
\SetKw{Null}{Null}
\SetKw{Return}{Return}
\SetKw{in}{in}
\SetKw{is}{is}
\SetKw{not}{not}
\SetKw{from}{from}
\SetKw{to}{to}
\SetKw{continue}{continue}
\SetKw{break}{break}

\alg{\falsify{}}{  
{\color{brown}\tcp{A DOMAIN-INDEPENDENT PROCEDURE}}

$tree \,\leftarrow\, $ []\\

$random\_meta\_state \,\leftarrow\, $\sampleMetaState{}\\
$root\_node \,\leftarrow\, Node(meta\_state\leftarrow random\_meta\_state, parent\leftarrow \Null)$\\
$tree.\mathtt{add}(root\_node)$

\While{\not timeout}{
$selection\_result \,\leftarrow\, \selectNode{tree}$

\uIf{$selection\_result$ \is $MetaState$}{
\Return $selection\_result$ {\color{teal}\tcp*{GOAL FOUND WHILE SAMPLING FOR NODE SELECTION!}}
}
\uElse{
$selected\_node \,\leftarrow\, selection\_result$

$expansion\_result \,\leftarrow\, selected\_node.\expandNode{tree,selected\_node}$\\
\uIf{$expansion\_result$ \is $MetaState$}{
\Return $expansion\_result$ {\color{teal}\tcp*{GOAL FOUND DURING NODE EXPANSION!}}
}
}
}
\Return \Null  {\color{teal}\tcp*{FAILED TO FIND FALSIFYING EXAMPLE---PERHAPS RESTART PROCESS}}

}

\BlankLine
\BlankLine

\proc{\sampleMetaState{}}{  
{\color{brown}\tcp{A DOMAIN-INDEPENDENT PROCEDURE}}

$random\_env \,\leftarrow\, \sampleenv()$\\
$initial\_scene \,\leftarrow\, Scene(random\_env, INITIAL\_STATE)$\\
$trajectory, history \,\leftarrow\, $\simulate{$initial\_scene$}\\
\Return $(random\_env, trajectory, history)$
}
\BlankLine
\BlankLine

\proc{\sampleenv{}}{  
{\color{brown}\tcp{A DOMAIN-SPECIFIC PROCEDURE}}

$env \,\leftarrow\, $ initialize environment of type ``track''\\
set track parameters\\
\For{i \from 1 \to number of obstacles}{
$(x,y,r) \,\leftarrow\, $ sample a random obstacle position\\
$env.obstacles.\mathtt{add}((x,y,r))$
}
\Return $env$

}

\BlankLine
\BlankLine

\caption{Meta-planning-based falsification using sampling-based forward search.}
\label{alg:fals}
\end{algorithm}

\begin{algorithm}[ht]
\scriptsize
\SetAlgoLined\DontPrintSemicolon
\setstretch{1.1}

\SetKwProg{alg}{Algorithm}{}{}
\SetKwProg{proc}{Procedure}{}{}

\SetKwFunction{selectNode}{select\_node}
\SetKwFunction{expandNode}{expand\_node}
\SetKwFunction{dtf}{distance\_to\_failure}
\SetKwFunction{perturbEnv}{perturb\_env\_elements}
\SetKwFunction{replaceEnv}{replace\_env\_elements}
\SetKwFunction{sampleStartScene}{sample\_start\_scene}
\SetKwFunction{sampleEndScene}{sample\_start\_scene}
\SetKwFunction{sampleMetaState}{sample\_meta\_state}
\SetKwFunction{envDist}{env\_distance}
\SetKwFunction{trajDist}{agent\_traj\_distance}
\SetKwFunction{sceneTrajDist}{scene\_traj\_distance}
\SetKwFunction{metaAction}{random\_meta\_control}
\SetKwFunction{simulate}{simulate\_system}
\SetKwFunction{metastatedist}{meta\_state\_distance}
\SetKwFunction{envdist}{env\_distance}
\SetKwFunction{sampleenv}{sample\_env}

\SetKw{True}{True}
\SetKw{False}{False}
\SetKw{Null}{Null}
\SetKw{Return}{Return}
\SetKw{in}{in}
\SetKw{is}{is}
\SetKw{not}{not}
\SetKw{from}{from}
\SetKw{to}{to}
\SetKw{continue}{continue}
\SetKw{break}{break}

\proc{\selectNode{$tree$}}{  
{\color{brown}\tcp{A DOMAIN-INDEPENDENT PROCEDURE}}

\uIf{{\color{purple} RANDOM\_NODE\_SELECTION}}
{
$ selected\_node \,\leftarrow\, $randomly selected node from $tree$\\
}
\uElseIf{{\color{purple} GREEDY\_NODE\_SELECTION}}{

$ selected\_node \,\leftarrow\, \argmax_{node \,\in\, tree} \dtf{node.meta\_state}$
}
\uElseIf{{\color{purple} RRT\_NODE\_SELECTION}}{
$r \,\leftarrow\, $ random number in the range $(0,1)$

\uIf{$r < {\color{purple} GOAL\_BIAS}$}{
$ selected\_node \,\leftarrow\, \argmax_{node \,\in\, tree} \dtf{node.meta\_state}$

}
\uElse{

\uIf{{\color{purple} STANDARD\_DISTANCE}}{

$random\_meta\_state \,\leftarrow\, $\sampleMetaState{}\\
\uIf{$\status(random\_meta\_state)=0$}{
\Return $random\_meta\_state$ {\color{teal}\tcp*{GOAL FOUND WHILE SAMPLING FOR NODE SELECTION!}}
}
\uElse{
$ selected\_node \,\leftarrow\, \argmax_{node \,\in\, tree} \metastatedist(node.meta\_state,random\_meta\_state)$
}
}
\uElseIf{{\color{purple} SIMPLIFIED\_DISTANCE}}{
$random\_env \,\leftarrow\, \sampleenv()$\\
$ selected\_node \,\leftarrow\, \argmax_{node \,\in\, tree} \envdist(node.meta\_state.env,random\_meta\_state.env)$
}

}
}
\Return $selected\_node$
}

\BlankLine
\BlankLine

\caption{Node selection.}
\label{alg:selection}
\end{algorithm}

\begin{algorithm}[ht]
\scriptsize
\SetAlgoLined\DontPrintSemicolon
\setstretch{1.1}

\SetKwProg{alg}{Algorithm}{}{}
\SetKwProg{proc}{Procedure}{}{}

\SetKwFunction{selectNode}{select\_node}
\SetKwFunction{expandNode}{expand\_node}
\SetKwFunction{dtf}{distance\_to\_failure}
\SetKwFunction{perturbEnv}{perturb\_env\_elements}
\SetKwFunction{replaceEnv}{replace\_env\_elements}
\SetKwFunction{sampleStartScene}{sample\_start\_scene}
\SetKwFunction{sampleEndScene}{sample\_start\_scene}
\SetKwFunction{sampleMetaState}{sample\_meta\_state}
\SetKwFunction{envDist}{env\_distance}
\SetKwFunction{trajDist}{agent\_traj\_distance}
\SetKwFunction{sceneTrajDist}{scene\_traj\_distance}
\SetKwFunction{metaAction}{random\_meta\_control}
\SetKwFunction{simulate}{simulate\_system}
\SetKwFunction{metastatedist}{meta\_state\_distance}
\SetKwFunction{envdist}{env\_distance}
\SetKwFunction{sampleenv}{sample\_env}
\SetKwFunction{findtimestamp}{find\_history\_compromise\_timestamp}
\SetKwFunction{mutate}{random\_replacement\_mutation}

\SetKw{True}{True}
\SetKw{False}{False}
\SetKw{Null}{Null}
\SetKw{Return}{Return}
\SetKw{in}{in}
\SetKw{is}{is}
\SetKw{not}{not}
\SetKw{from}{from}
\SetKw{to}{to}
\SetKw{continue}{continue}
\SetKw{break}{break}

\proc{\expandNode{$tree,node$ }}{  
{\color{brown}\tcp{A DOMAIN-INDEPENDENT PROCEDURE}}

\For{i \from 1 \to {\color{purple} EXPANSION\_BREADTH}}{
$new\_meta\_state \,\leftarrow\,$\metaAction{$node.meta\_state$}\\
$new\_node \,\leftarrow\, Node(meta\_state\leftarrow new\_meta\_state, parent\leftarrow node)$\\
$tree.\mathtt{add}(new\_node)$\\
\uIf{$\status(new\_meta\_state)=0$}{
\Return $new\_meta\_state$ {\color{teal}\tcp*{GOAL FOUND DURING NODE EXPANSION!}}
}

}
\Return \Null {\color{teal}\tcp*{GOAL NOT YET FOUND}}
}

\BlankLine
\BlankLine

\proc{\metaAction{$meta\_state$ }}{  
{\color{brown}\tcp{A DOMAIN-INDEPENDENT PROCEDURE (META-DYNAMICS)}}
$ mutated\_env \,\leftarrow\, $\mutate{$meta\_state.\env$}
{\color{teal}\tcp*{MUTATE ENV}}

{\color{teal}\tcp*{CALC TIMESTAMP IN WHICH MUTATION STARTS AFFECTING OBSERVATION HISTORY}}
$history\_compromise\_timestamp \,\leftarrow\, $\findtimestamp{$meta\_state,mutated\_env$} \\

$history\_comporomise\_state \,\leftarrow\, meta\_state.traj[history\_compromise\_timestamp]$\\
{\color{teal}\tcp*{SIMULATE STARTING FROM TIMESTAMP, AND INCREMENTALLY UPDATE ORIGINAL TRAJECTORY/HISTORY}}

$initial\_scene \,\leftarrow\, Scene(mutated\_env,history\_comporomise\_state)$\\
$trajectory\_suffix, history\_suffix \,\leftarrow\, $\simulate{$initial\_scene$}\\

$ updated\_traj \,\leftarrow\, [ meta\_state.traj[0:history\_compromise\_timestamp], trajectory\_suffix$]\\

$ updated\_history \,\leftarrow\, [ meta\_state.history[0:history\_compromise\_timestamp], history\_suffix$]\\

\Return $(mutated\_env,updated\_traj,updated\_history)$
{\color{teal}\tcp*{RETURN UPDATED META STATE}}

}

\BlankLine
\BlankLine

\proc{\mutate{$env$}}{  
{\color{brown}\tcp{A DOMAIN-SPECIFIC PROCEDURE (THE $\omp$ OPERATOR)}}
$ mutated\_env \,\leftarrow\, copy(\env)$\\

\uIf(\tcp*[f]{{\color{teal}DETERMINE ENV MUTATION SIZE}}){{\color{purple} CONST\_MUTATION\_WIDTH}}{
$n \,\leftarrow\,$STEP\_SIZE
}
\uElseIf{{\color{purple} RANDOM\_MUTATION\_WIDTH}}{
$n \,\leftarrow\, $randomly selected number from $1$ to $\mathtt{len}(mutated\_env.obstacles)$ of obstacles to replace
}
{\color{teal}\tcp*{PERFORM RANDOM ENV MUTATION}}

$obstacles\_to\_replace \,\leftarrow\, $randomly selected $n$ obstacles from $mutated\_env.obstacles$

\For{$obstacle$ \in $obstacles\_to\_replace$}{

$mutated\_env.obstacles.\mathtt{remove}(obstacle)$\\

\uIf{{\color{purple} UNLIMITED\_MUTATION\_DEPTH}}
{
$new\_obstacle \,\leftarrow\, $ uniformly sample a random obstacle position\\
}

\uElseIf{{\color{purple} LIMITED\_MUTATION\_DEPTH}}{

$new\_x \,\leftarrow\, obstacle.x + $ randomly sampled scalar noise\\
$new\_y \,\leftarrow\, obstacle.y + $ randomly sampled scalar noise\\
$new\_r \,\leftarrow\, obstacle.r + $ randomly sampled scalar noise\\
$new\_obstacle \,\leftarrow\, (new\_x,new\_y,new\_r)$
}
$mutated\_env.obstacles.\mathtt{add}(new\_obstacle)$\\

}
\Return $mutated\_env$
}

\BlankLine
\BlankLine

\proc{\findtimestamp{$meta\_state, new\_\env$}}{
{\color{brown}\tcp{A DOMAIN-INDEPENDENT VERSION}}

\For{$i$ \from $0$ \to $\texttt{length}(meta\_state.traj)-1$}{
$new\_\obs \,\leftarrow\, h(meta\_state.traj[i],new\_env)$  {\color{teal}\tcp*{SIMULATE OBSERVATION IN NEW ENV}}

\uIf{\not $new\_\obs = meta\_state.observation\_history[i]$}{
\Return i {\color{teal}\tcp*{OBSERVATION INVALID IN NEW ENV}}
}

}
\Return $\texttt{length}(meta\_state.traj)$  {\color{teal}\tcp*{HISTORY VALID IN NEW ENV}}

}

\BlankLine
\BlankLine

\proc{\findtimestamp{$meta\_state, new\_\env$}}{
{\color{brown}\tcp{AN EFFICIENT DOMAIN-SPECIFIC VERSION: AVOIDING SIMULATION USING KNOWLEDGE ON SENSOR MODEL}}

\For{$i$ \from $0$ \to $\texttt{length}(meta\_state.traj)-1$}{
$observed\_area \,\leftarrow\, $ estimate observed portion of track in $meta\_state.observation\_history[i]$ based on sensor properties\\
\uIf{any of added/removed obstacles overlap with $observed\_area$}{
\Return i {\color{teal}\tcp*{OBSERVATION INVALID IN NEW ENV}}
}

}
\Return $\texttt{length}(meta\_state.traj)$  {\color{teal}\tcp*{HISTORY VALID IN NEW ENV}}

}

\BlankLine
\BlankLine

\caption{Node expansion.}
\label{alg:expansion}
\end{algorithm}

\newpage $ $ \newpage
    \section{Theoretical interpretation and discussion}
\label{appen:theory}

In the challenging example we examined, the environment-state specified the location for multiple obstacles, and the car only failed when these obstacles were in specific and highly-correlated configurations.
Also, small changes in the location of obstacles could cause discrete changes in the homotopy class of the car trajectory, resulting in large changes in its distance-to-failure. 
This overall meant that the search goal corresponded to many small and sparsely-scattered goal meta-regions, and that the optimization field was non-smooth and non-convex, with many local minima.

\begin{figure}[ht]
\centering
\includegraphics[width=0.5\textwidth]{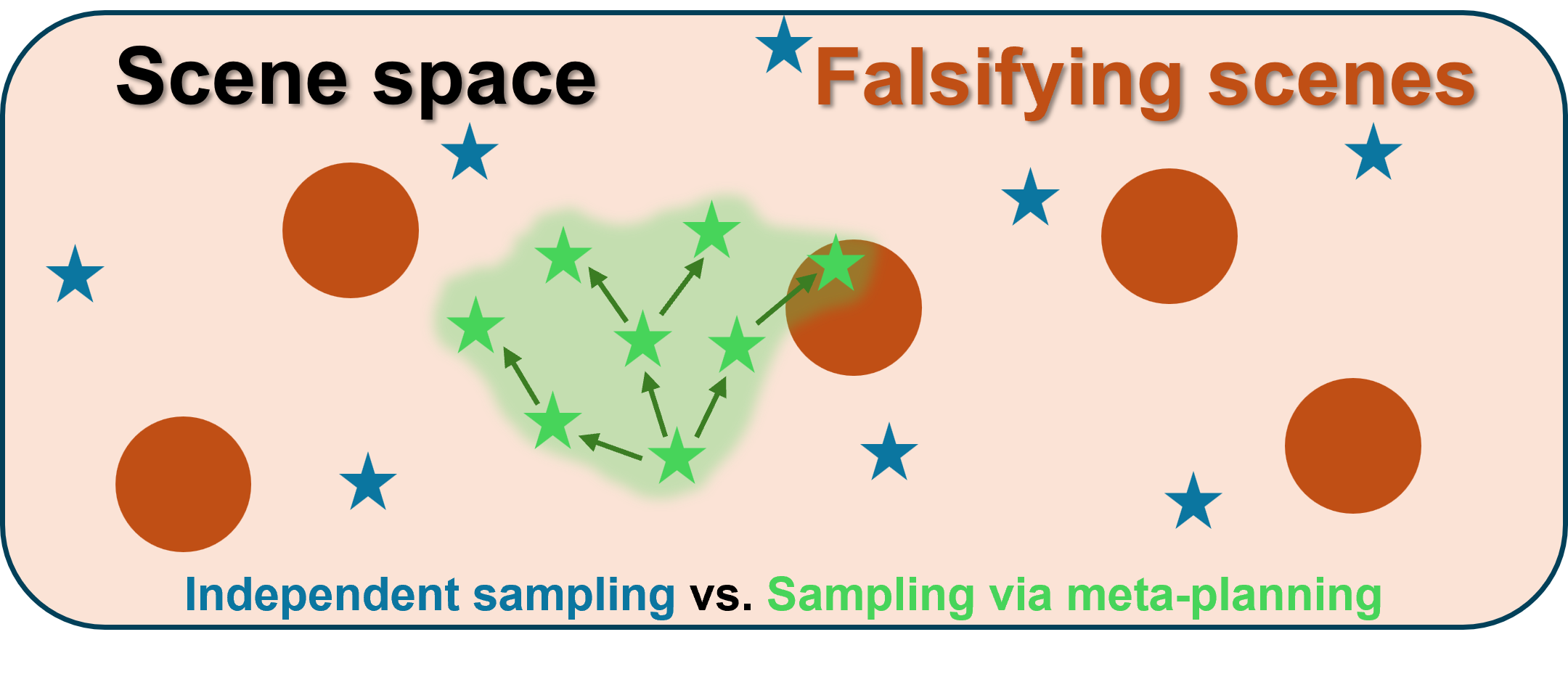}
	\caption{Illustration of the scene-sampling patterns of our incremental, tree-based, meta-planning sampler (in green), and a standard, sequential sampler (in blue). In this challenging case, where the search goals are sparse, our meta-planning approach seem to require less sample until ``hitting'' one of the goal regions. }
\label{fig:pattern}
\end{figure}

One of the major benefits of our meta-planning, in comparison to the other approaches mentioned, is that it expresses an \emph{incremental generative model} for system trajectories. Meaning, each new system trajectory generated for examination is not a free-standing sample, but is generated directly from an existing one. This property, together with the choice to grow a tree of samples, and not a simple chain, results in a sampling pattern that densely covers whole regions of the search space.
Further, since our tree is not fully-greedy, a region can grow in various directions. Allowing for mutations of high-width, as we mentioned, also allow us to ``re-root the tree,'' essentially growing multiple of such dense regions in different parts of the space, ensuring proper coverage and exploration. 
Overall, this sampling scheme steadily and effectively ``combs'' through the environment space until finding a falsifying example.
This pattern seems to be very effective in challenging scenarios like ours, when the goals are sparse (as illustrated in \cref{fig:pattern}). In these cases, standard sequential-sampling approaches, which result in a more sparse pattern , may struggle to to ``hit'' a small goal region.

It is also likely that the complex relationship between the input and output, as we described above, particularly debilitated the ability of the BO-baseline to learn a useful model to guide its sampling.
Another point which may have caused this struggle is the arbitrary fixation of the obstacles into specific variables (``obstacle 1/2/3''), which does not capture the inherent symmetry under permutation of the obstacle collection.

\paragraph{Meta-planning vs. genetic algorithms.}
While both meta-planning and the ``generation-based'' genetic algorithm we examined utilized mutation operations for incremental sample generation, as proven by our experiments, meta-planning was a more effective search approach. 
This comes due to numerous differences between the approaches:
first and most importantly, meta-planning performs search in the composite space of inputs $\times$ outputs, while genetic algorithms search only the input space.
Further, unlike genetic algorithms, when solving the meta-planning problem, we are not forced to perform serial ``blanket'' updates to the entire sample population, but use evolutionary operations to grow a planning-tree of samples, by dynamically selecting and evolving a single promising sample at a time. By such, we are also not restricted to sample population of a predefined size, as the number of leaves in the planning tree can dynamically grow, according to the number of promising search directions discovered. It is also important to note that evolution operators can cause the ``value'' of a sample to decrease; yet, while those algorithms only maintain the latest generation of samples, our tree-based structure allows us to backtrack during the search to revisit past samples, if those seemed more promising than the latest generation. Maintaining all the samples also allows us to leverage space-coverage heuristic, to make sure the space is properly explored, and the samples do not ``collapse'' to a local minimum.
We may also note that in our specific example, the crossover operation used by the genetic algorithm might not have been effective, as combining two ``halves'' of close-to-failure environments did not necessarily result in a close-to-failure environment.
All-in-all, all of these properties proved to lead to better falsification performance of the meta-planning approach. 
In fact, we can even view generation-based genetic algorithms as a restricted subset of the solution algorithms we may use to solve our meta-planning formulation. For example, it is easy to see that the behavior of a standard evolutionary algorithm can be achieved with a redundant meta-planning forward-search algorithm, by first extending an initial set of samples from a root node, and then continuing to grow the tree using a Breadth-First-Search (BFS) strategy. More advanced variations can be demonstrated as well, by appropriately managing the node priority-queue.

\end{document}